\documentclass[twoside]{article}

\usepackage[preprint]{aistats2027}

\usepackage[round]{natbib}

\usepackage{amsmath}
\usepackage{amssymb}
\usepackage{amsthm}
\usepackage{thmtools}  %
\usepackage{graphicx}
\usepackage{tikz}
\usepackage{booktabs}
\usepackage{placeins}   %
\usepackage{xcolor}
\usepackage{algorithm}
\usepackage{algpseudocode}
\usepackage{xspace}
\definecolor{linknavy}{RGB}{0,82,163}      %
\usepackage{xurl}        %
\usepackage[colorlinks=true,allcolors=linknavy]{hyperref}

\DeclareMathOperator*{\E}{\mathbb{E}}
\DeclareMathOperator*{\argmin}{arg\,min}
\DeclareMathOperator*{\argmax}{arg\,max}

\DeclareMathOperator*{\conv}{conv}  %

\newcommand{\reals}{\mathbb{R}}
\newcommand{\domain}{\mathcal{X}}  %
\newcommand{\ind}[1]{\mathbb{I}{[#1]}}
\newcommand{\simpleF}{\mathcal{F}}  %
\newcommand{\simpleG}{\mathcal{G}}  %
\newcommand{\cell}{\mathcal{A}}            %
\newcommand{\partition}{\mathcal{P}}       %
\newcommand{\partitionclass}{\mathcal{C}}  %
\newcommand{\trees}{\mathcal{T}}    %
\newcommand{\leaves}{\mathcal{L}}   %
\newcommand{\leaf}{\cell}           %
\newcommand{\confusionmatrix}{\mathbf{p}^\leaf} %

\newcommand{\half}{\frac{1}{2}}

\newcommand{\origbranches}{\texttt{Branches}\xspace}
\newcommand{\branches}{\texttt{PI-Branches}\xspace}
\newcommand{\convert}{\texttt{CART-and-convert}\xspace}
\newcommand{\cart}{\texttt{CART}\xspace}
\newcommand{\pimcart}{\texttt{PI-CART}\xspace}

\newcommand{\defer}{\bot}
\newcommand{\lzo}{\ell_{0/1}}
\newcommand{\lpim}{\ell_{\textrm{pim}}}
\newcommand{\llog}{\ell_{\textrm{log}}}
\newcommand{\lpimlog}{\ell_{\textrm{pim}}^{\log}}

\newcommand{\pimentropy}{H_{\textrm{pim}}}

\newtheorem{theorem}{Theorem}

\definecolor{pimteal}{RGB}{31,111,107}
\newcommand{\pimdiff}[1]{\textcolor{pimteal}{\ifmmode\boldsymbol{#1}\else{\boldmath\bfseries #1}\fi}}

\makeatletter
\newenvironment{breakablealgorithm}
  {\begin{center}\refstepcounter{algorithm}%
   \hrule height.8pt depth0pt \kern2pt%
   \renewcommand{\caption}[2][\relax]{%
     {\raggedright\textbf{\ALG@name~\thealgorithm}\ ##2\par}%
     \kern2pt\hrule\kern2pt}}
  {\kern2pt\hrule\relax\end{center}}
\makeatother

\begin{document}

\runningtitle{Partially Interpretable Decision Trees}
\runningauthor{Ludovic Schwartz, Federico Berton, Tim van Erven}

\twocolumn[

\aistatstitle{Partially Interpretable Decision Trees}

\aistatsauthor{ Ludovic Schwartz \And Federico Berton \And Tim van Erven }

\aistatsaddress{ University of Amsterdam \And University of Amsterdam \And University of Amsterdam } ]

\begin{abstract}
Partially interpretable models extend simple interpretable
models with the option to defer complicated inputs to an accurate but
uninterpretable black-box model.
By deferring more inputs, the model
gets more accurate; by deferring fewer, it gets more interpretable.

Existing procedures to train such partially interpretable models build
on linear models or decision rules. We propose three algorithms for
decision trees: a partially interpretable extension of the existing
\origbranches algorithm, which performs exact search to optimize the
tree; a conversion algorithm that we run on top of standard \cart, which
is computationally more efficient; and a generalized \cart whose
splitting criterion takes into account the option to defer while growing
the tree.

Perhaps surprisingly, the performance of the three algorithms is similar
across many real-world datasets. We can explain this through a formal
 bound on the potential gain of smart tree construction, provided the black box accuracy is sufficiently homogeneous across the
domain. We also construct a synthetic dataset where the accuracy of the
black box is deliberately inhomogeneous. There, our deferral-aware
generalization of \cart does outperform regular \cart and matches the
performance of exact search.

Finally, we evaluate an
application to
concept bottleneck models. Partially interpretable decision trees can replace uninterpretable neural networks after the bottleneck.
On undeferred inputs, the model is fully interpretable, and deferring preserves accuracy.

\end{abstract}

\section{Introduction}
\label{sec:intro}

When selecting machine learning models for classification, there is a
tension between interpretability and accuracy. Certain simple models,
such as linear classifiers, rule lists, and decision trees
\citep{Breim_F_O_S84Classificationregressiontrees,
Rives_87Learningdecisionlists,
Angel_L_A_S_R17Learningcertifiablyoptimal} are inherently interpretable
(provided they are of reasonable size), but this is only possible
because they impose strong restrictions on the class of functions they
can express, and as a consequence they cannot learn complicated
relations. Conversely, richer function classes like neural networks or
support vector machines can, for instance, approximate all
continuous functions on a compact domain
\citep{HofmannScholkopfSmola2008,HornikStinchcombeWhite1989}, but are
not inherently interpretable and behave as a black box.

A commonly proposed solution is to use simple, interpretable models to
locally approximate accurate, more complicated black box models, and use
the local approximation as a way to interpret the black box. For
instance, \texttt{LIME} \citep{RibeiroSinghGuestrin2016LIME},
\texttt{SHAP} \citep{Lundb_L17unifiedapproachinterpreting},
\texttt{SmoothGrad} \citep{SmilkovThoratKimEtAl2017} and
\texttt{Integrated Gradients} \citep{SundararajanTalyYan2017} may all be
viewed as local linear approximations \citep{HanSrinivasLakkaraju2022}.
Unfortunately, a major limitation of these local approximations is that
their explanations provide no guarantees: we typically do not know how
accurate the local approximation is, or over which region it is valid.
Therefore users cannot base any decisions on the local approximations
with any certainty.

To resolve this, we need explanations that are \emph{sufficient}: the
explanation determines the prediction and points sharing the same
explanation will have the same prediction
\citep{Frost_L_M_M24Partiallyinterpretablemodels}. Sufficiency is
satisfied by so-called partially interpretable models, which consist of
a hybrid between an interpretable simple model and an accurate black-box
model \citep{Wang_L21Hybridpredictivemodels,
Frost_L_M_M24Partiallyinterpretablemodels}. On simple inputs, the
interpretable model will make predictions itself, which are therefore
interpretable, and on complicated inputs it will defer to the black box
for prediction. Such hybrid models offer a trade-off between accuracy
and interpretability: by deferring more inputs, the hybrid model gets
more accurate, and by deferring fewer it gets more interpretable.
Importantly, the decision whether to defer an input is always handled by
the interpretable model, and is therefore also interpretable.

Training procedures for partially interpretable models have so far been
developed when the simple model is linear
\citep{Wang_L21Hybridpredictivemodels} or consists of a list of decision
rules
\citep{Wang_L21Hybridpredictivemodels,Frost_L_M_M24Partiallyinterpretablemodels,FerryLabergeAivodji2024},
but
not yet for the important class of decision trees. How to learn
partially interpretable trees is the topic of this paper.

\paragraph{New algorithms}

First, in Section~\ref{sec:decision-tree}, we introduce three algorithms
to learn partially interpretable decision trees. The first, \branches,
is a partially interpretable extension of the existing \origbranches
algorithm of \citet{Chaou_R_B25BranchesEfficientlyseeking}, which
performs exact search\footnote{The search may not always be fully exact.
See Section~\ref{sec:branches} for details.} to optimize the tree. Since
exact search is NP-complete already without deferring
\citep{LaurentRivest1976}, this is computationally feasible only when
the number of features is not too large. The second, \convert, takes the
tree learned by the computationally efficient \cart algorithm
\citep{Breim_F_O_S84Classificationregressiontrees}, and makes it defer
on a subset of its leaves. The third, \pimcart, is also based on \cart,
but generalizes it by changing its splitting criterion to take into
account the option to defer already while growing the tree.

\paragraph{A theoretical insight}

We originally intended \convert as a simple baseline, but it performs
surprisingly well in experiments. In Section~\ref{sec:conversion_bound}
we provide a theoretical explanation: we compare the conversion
procedure to an oracle that knows the optimal tree, and bound the gap
between the corresponding curves of optimal trade-offs between expected
error and proportion of inputs that are deferred.
Theorem~\ref{thm:one_eight_bound} shows that these curves can differ by
at most $\frac{1}{8} (1 + 2 \epsilon)^2 + \delta + \alpha$, where
$\epsilon$ measures the suboptimality of \cart compared to the best tree
that never defers, $\delta$ measures the suboptimality of the conversion
procedure and $\alpha$ measures how much the performance of the black
box varies across the domain. \citet{Kearn_M99boostingabilitytopdown}
show that \cart can drive $\epsilon$ to $0$ in the PAC learning setting,
and we can expect $\delta$ to go to $0$ as the sample size increases.
Therefore, if the inhomogeneity $\alpha$ is also small, the conversion
procedure's performance is never much more than $1/8 = 12.5\%$ worse
than the oracle. The theorem further provides a problem-dependent bound
that is significantly tighter if there exists a tree that can achieve
small expected error without deferring. This shows that the potential
gains from more sophisticated tree growing procedures are limited.

\paragraph{Experiments}

In Section~\ref{sec:experiments} we provide experiments to compare the
different tree learning algorithms, which show they tend to have similar
performance except in special cases. On real data, the gains of
\branches and \pimcart over \convert are small and inconsistent: over
$15$ real data sets, the three methods lie within $0.60\%$ error of one
another on average. In fact, on $10$ of these data sets the difference is within
$1.2\%$ error. There are larger gaps, of up to $10.9\%$ error on several
small datasets, but they do not consistently favor any single method.
Inspired by Theorem~\ref{thm:one_eight_bound}, we further construct a
synthetic data set on which the performance of the black box is
deliberately inhomogeneous (see Appendix~\ref{app:exp:synthetic}). In
this case we do observe that \branches and \pimcart are better able to
adapt the tree structure to take this into account, and outperform
\convert.

\paragraph{Application to concept bottleneck models}

Finally, we combine our methods with concept bottleneck models
\citep{Koh_N_T_M_P_K_L20Conceptbottleneckmodels}, which predict in two
stages: a (potentially complicated) map from raw features to a layer
of human-understandable concepts, followed by an interpretable model $f$
that maps concepts to predictions. Even though $f$ is supposed to be
interpretable, it is common in the literature to use a two-layer
feedforward network. Instead, we let $f$ be a partially interpretable
decision tree that can defer to a black box that predicts labels from
raw features directly. This has the advantage that $f$ is truly
interpretable on the inputs on which it does not defer. We find that, on
CUB-200 data, a depth-$8$ tree over hard concepts matches the accuracy of a soft concept
bottleneck model with 60k parameters for the network $f$, while
providing sufficient explanations on $57\%$ of the data.

\paragraph{Further related work}

Our setup is formally similar to \emph{learning to reject}, in which
binary classification is extended with the option to reject an instance
at a fixed cost $c \in (0,1/2)$ instead of predicting a label
\citep{Chow1970,BartlettWegkamp2008,CortesDeSalvoMohri2016}, and its
generalization \emph{learning to defer}, where predictions may be
deferred to a human expert
\citep{MadrasPitassiZemel2018,MozannarSontag2020}. Variants of these
settings correspond to what we will call the Lagrangian problem
formulation in Section~\ref{sec:CPvsLP}. The motivation is very
different, however, because we require interpretability of the model and
rejection function, whereas the goal of learning to defer is to reduce
classification mistakes in uncertain situations, and interpretability
plays no role.

A variant of partially interpretable models is also developed by
\citet{PanWangHara2020}, who allow the user to choose between a simple
model and the black box.
\citet{CravenShavlik1994,CravenShavlik1995,LakkarajuKamarCaruanaLeskovec2017}
approximate a black box model by decision rules or trees, and
\citet{Bress_C_E_M_M_T24theoryinterpretableapproximations} characterize
theoretically when black box models can be approximated by trees.

\section{Partially Interpretable Models}
\label{sec:pim}

\subsection{Setting}

We first consider the standard binary classification setting. Let $S =
\big((X_1,Y_1), \ldots, (X_n,Y_n)\big)$ be a sample of independent draws
from a common distribution $P$, where $X_i$ takes values in $\domain
\subseteq \reals^d$ and $Y_i$ is a label in $\{0,1\}$. Given a
prediction $\hat y$ and outcome $y$ in $\{0,1\}$, let $\lzo(y,\hat
y) = \ind{\hat y \neq y}$ denote the $0/1$-loss. For a given classifier
$f : \domain \to \{0,1\}$, let
\begin{align*}
  R(f) &= \E_{(X,Y) \sim P}[\ell_{0/1}(Y,f(X))] = P\big(f(X) \neq Y\big), \text{ and}\\
  R_n(f) &= ~\frac{1}{n} \sum_{i=1}^n \ell_{0/1}(Y_i,f(X_i))
\end{align*}
denote the risk and the empirical risk, respectively. Any classifier
$f^* \in \argmin R(f)$ is called Bayes-optimal.

To extend this setting to allow for partially interpretable models,
suppose we have access to a high-quality reference classifier $b$, which
is considered an uninterpretable black box. For instance, $b$ might be
the output of boosting, a random forest or a neural network. Then we can
allow for an \emph{extended prediction} $\hat y \in \{0,1,\bot\}$, where
we take $\bot$ to mean ``defer the prediction to $b$''. Given an
\emph{extended classifier} $f : \domain \to \{0,1,\bot\}$ that makes
such extended predictions, let $f_{|b}$ denote the induced classifier that
predicts
\[
  f_{|b}(x) =
  \begin{cases}
    f(x) & \text{if $f(x) \in \{0,1\}$}\\
    b(x) & \text{if $f(x) = \bot$}.
  \end{cases}
\]
We will be interested in extended classifiers $f$ that are simple enough
to be interpretable (e.g.\ decision trees, decision rules, linear
classifiers), so that $f_{|b}$ is interpretable on inputs $x$ for which
$f(x) \in \{0,1\}$. When $b$ is highly accurate, such an extended
classifier can reduce its risk $R(f_{|b})$ by deferring more inputs
to~$b$, at the cost of increasing the proportion of inputs it leaves
uncovered, which we will call the (empirical) \emph{uncoverage}:
\begin{align*}
  U(f) &= P\big(f(X) = \bot\big),
       &
  U_n(f) &= \frac{1}{n} \sum_{i=1}^n \ind{f(X_i) = \bot}.
\end{align*}

\subsection{Constrained versus Lagrangian Optimization}
\label{sec:CPvsLP}

There are two natural ways to formulate the trade-off between minimizing
the empirical risk and the empirical uncoverage over a class of
interpretable extended classifiers $\simpleF$: the \emph{constrained
problem}
\begin{align*}
  \min_{f \in \simpleF} ~R_n(f_{|b}) &
  & \text{subject to } U_n(f) \leq t,
  \tag{CP}
\end{align*}
parametrized by $t \geq 0$, and the \emph{Lagrangian problem}
\begin{equation*}
  \min_{f \in \simpleF} ~ \Big\{R_n(f_{|b}) + \lambda U_n(f)\Big\},
  \tag{LP}
\end{equation*}
for $\lambda \geq 0$. The difference is that the CP sets a hard
constraint $t$ on the uncoverage, whereas the LP introduces a soft cost
$\lambda$ for uncovered inputs.

The CP is the most general formulation, because, by varying $t$, its
solutions trace out the whole Pareto frontier of possible optimal
trade-offs between risk and uncoverage. We can recover some of these
optimal trade-offs by solving the LP with varying $\lambda$ as well,
because any solution $\hat f$ to the LP is also a solution to the CP for
$t = U_n(\hat f)$, and therefore also sits on the Pareto frontier. This is
attractive, because the LP is often computationally easier, but
unfortunately it can miss some Pareto optimal trade-offs that do not sit
on the lower convex hull of Pareto optimal trade-offs (because the
optimization problem is not convex, and therefore has potentially
non-zero duality gap). This is discussed further in
Appendix~\ref{app:sec:pim}, which also shows that the gap can be
substantial: the appendix exhibits a data set of $n = 7$ points on which
a Pareto optimal trade-off is not the solution to an LP for any
$\lambda$, and extends it to a family of data sets on which the
resulting suboptimality approaches $\tfrac14$.

\section{Learning Partially Interpretable Trees}
\label{sec:decision-tree}

In this section we discuss three algorithms for learning partially
interpretable models when the class of interpretable extended
classifiers $\simpleF$ consists of binary decision trees $\trees$. We
will represent a binary tree $T = (\leaves,\sigma)$ by the partition
$\leaves = \{\leaf_1,\ldots,\leaf_{|\leaves|}\}$ of $\domain$ that it
induces, where each partition cell $\leaf \in \leaves$ is a leaf of the
tree, and $\sigma : \leaves \to \{0,1,\bot\}$ is an \emph{extended
labeling function} that assigns a label $\sigma(\leaf)$ to each leaf
$\leaf$, including the option $\bot$ to defer. Letting $\leaf_\leaves(x)
\in \leaves$ denote the leaf that contains $x$, we then also write $T(x)
= \sigma\big(\leaf_\leaves(x)\big)$ for the prediction of the tree on
input $x$. Each internal node splits on a threshold of a single feature (axis-aligned splits).

\subsection{Exact Search with Partially Interpretable Branches}
\label{sec:branches}

The \origbranches{} algorithm of \citet{Chaou_R_B25BranchesEfficientlyseeking} performs exact search to
minimize $ \big\{ R_n(T_{|b}) + \gamma |\leaves| \big\}$
over the class of trees that never defer. Its hyperparameter $\gamma
\geq 0$ trades off between the number of leaves and the fit on the data.
It further requires binary features, and takes an optional additional
parameter to restrict the maximum depth of the trees, in which case the
search is only approximately exact.

We extend this to a partially interpretable variant \branches{} that
allows deferring and takes the uncoverage into account by minimizing
\begin{equation}\label{eqn:pi_branches_objective}
  \min_{T = (\leaves,\sigma)} ~
  \Big\{ R_n(T_{|b}) + \lambda U_n(T) + \gamma |\leaves|\Big\}.
\end{equation}

\origbranches is an AO*-type algorithm performing Best-First-Search within an AND/OR graph
search framework. The only modification \branches requires is in how a terminal state (a leaf) is scored.

For a \emph{branch} $l$ (a candidate root-to-leaf path), let $n(l)$ be the number of training points that follow it,
$n_{k^*}(l)$ those in the majority class, and $n_b(l)$ those the black box predicts correctly. \origbranches scores a
leaf by its errors $n(l) - n_{k^*}(l)$; \branches replaces this by the
minimum of the
\origbranches score and $n(l) - n_b(l) + \lambda\, n(l)$, which is the
cost of deferring.
The search heuristic is also adjusted accordingly. See
Appendix~\ref{app:Branches} for a complete description of the algorithm.

\subsection{CART and Convert}
\label{sec:cart-convert}

The regular \cart algorithm \citep{Breim_F_O_S84Classificationregressiontrees} learns a decision tree $\hat T =
(\hat \leaves,\hat \sigma)$ without deferring, so with $\hat \sigma :
\hat \leaves \to \{0,1\}$. A simple, but surprisingly effective approach to
turn this into a partially interpretable decision tree is to fix the
tree structure and hence the leaves $\hat \leaves$, but replace $\hat
\sigma$ by an extended labeling function that optimizes either the
constrained problem or the Lagrangian problem. We then obtain the
following two variants.

\paragraph{Convert for constrained problem}

Given $\hat \leaves$, optimize
\begin{align*}
  \min_{\sigma} ~R_n\big((\hat \leaves,\sigma)_{|b}\big) &
  & \text{subject to } U_n\big((\hat \leaves,\sigma)\big) \leq t.
\end{align*}
If we do not defer on a leaf, then the optimal label $\sigma(\leaf) =
\argmin_{\hat y \in \{0,1\}} \sum_{i: X_i \in \leaf} \ell_{0/1}(Y_i,\hat
y)$ is the majority class among the data points in that leaf, so the
remaining decision is when to defer. This gives rise to a knapsack
problem, where the leaves are the items
\citep{KellererPferschyPisinger2004} and a leaf is included in the
knapsack when the tree defers on that leaf. (See
Appendix~\ref{app:knapsack} for details.)

\paragraph{Convert for Lagrangian problem}

Given $\hat \leaves$, optimize
\begin{equation}\label{eqn:CARTLagrangianConvert}
  \min_{\sigma} ~\Big\{R_n\big((\hat \leaves,\sigma)_{|b}\big) + \lambda
  U_n\big((\hat \leaves,\sigma)\big)\Big\}.
\end{equation}
This formulation is easy to solve, because it makes the choice to defer
independent between leaves: the optimal choice is to defer for leaf
$\leaf$ whenever the loss of deferring, penalized by $\lambda$, is less
than the loss of not deferring:
\[
  \sum_{i:X_i \in \leaf} \big(\ell_{0/1}(Y_i,b(X_i)) +
  \lambda\big)
  <
  \min_{\hat y \in \{0,1\}} \sum_{i:X_i \in \leaf} \ell_{0/1}(Y_i,\hat y).
\]
In Appendix~\ref{app:exp:convert}, we experimentally compare the two methods. We measure the
difference in aggregate score (defined in Section~\ref{sec:exp:trackb}) between the two conversion approaches,
and find it below $0.3$ points (Table~\ref{tab:convert-rules}).

To limit overfitting, we prune the grown trees by cost-complexity
pruning \citep{Breim_F_O_S84Classificationregressiontrees}, which is
equivalent to minimizing the \branches objective
\eqref{eqn:pi_branches_objective} over all subtrees of the grown tree.

\subsection{Partially Interpretable CART}
\label{sec:pimcart}
\convert provides a straightforward way to train partially interpretable trees. However, information about the black box
$ b $
is not used when growing the tree. A natural extension is to consider a splitting criterion that incorporates
information about $ b $.

To do so, we start from the entropy criterion of \cart. If a fraction $m$ of the $n_\leaf = |\{i : X_i \in \leaf\}|$
training points in a leaf $\leaf$ are positive, the entropy of $\leaf$ is $H(m) = -m \log m - (1-m) \log (1-m)$. The
entropy has a useful interpretation in terms of soft predictions. Suppose that, instead of a single label, the leaf
predicts a probability distribution $\hat p \in \Delta(\{0,1\})$ over the two labels and that it is evaluated by the log-loss
$\llog(y, \hat p) = -\log \hat p(y)$. The best soft prediction is then $\hat p(1) = m$, and its average log-loss is
exactly the entropy:
\begin{equation*}
  H(m) = \min_{\hat p \in \Delta(\{0,1\})} \frac{1}{n_\leaf} \sum_{i : X_i \in \leaf} \llog(Y_i, \hat p).
\end{equation*}
This also explains why the entropy is a sensible splitting criterion: since $1 - p \le -\log p$, the average
log-loss of any $\hat p$ bounds its expected error $\frac{1}{n_\leaf}\sum_{i : X_i \in \leaf}(1 - \hat p(Y_i))$, so $H(m)$
is an upper bound on the error of the majority label, the smallest such error.

To extend the entropy criterion to partially interpretable decision trees, we want to mimic this construction with a
suitable generalization of the log-loss to soft predictions $\hat{p} \in \Delta(\{0,1,\defer\})$,
which takes the black box into account.
Given that deferred points are processed by the black box, the distribution $\hat{p}\in \Delta(\{0,1,\defer\})$ induces on
$\{0,1 \}$ the following distribution $\hat{p}_x(y) = \hat{p}(y) + \hat{p}(\defer)\,\ind{b(x) = y}$.
This motivates the following definition of the PIM log-loss:
\begin{equation*}
  \lpimlog(x,y,\hat{p})
  =
  -\log\big(\hat{p}(y) + \hat{p}(\defer)\,\ind{b(x)=y}\big)
  + \lambda\,\hat{p}(\defer).
\end{equation*}

This loss upper-bounds the penalized $0/1$-loss
\begin{equation*}
  \lpim(x,y,\hat{y}) = \ind{\hat{y} \notin \{y, \defer\}} + \ind{\hat{y} = \defer} \big(\ind{b(x) \neq y} + \lambda\big)
\end{equation*}
of a prediction $\hat y$ drawn from $\hat p$. Indeed, its expected value is
$\mathbb{E}_{\hat y \sim \hat p}\,\lpim(x,y,\hat y) = 1 - \hat p_x(y) + \lambda\,\hat p(\defer)$, and $1 - p \le -\log p$
applied to $p = \hat p_x(y)$ gives $\mathbb{E}_{\hat y \sim \hat p}\,\lpim(x,y,\hat y) \le \lpimlog(x,y,\hat p)$.
Averaged over the leaf and minimized over $\hat p$, the left-hand side is the smaller of the error of the majority
label and the error of $b$ on $\leaf$ plus $\lambda$, which is the cost of the leaf under the deferral rule of
Section~\ref{sec:cart-convert}. The PIM-entropy defined below is therefore an upper bound on this cost, just as the
entropy bounds the error of the majority label.

Let $p_{j,k}^\leaf$ be the fraction of the points in $\leaf$ with $Y = j$ and $b(X) = k$, for $j, k \in \{0,1\}$,
collected in the confusion matrix $\confusionmatrix = (p_{j,k}^\leaf)$; in particular $m = p_{1,0}^\leaf + p_{1,1}^\leaf$.
The average PIM log-loss on $\leaf$ depends on the leaf only through $\confusionmatrix$, and we define the PIM-entropy
of $\leaf$ as its minimum:
\begin{equation*}
  \pimentropy(\confusionmatrix)
  :=
  \min_{\hat{p} \in \Delta(\{0,1,\defer\})}
    \frac{1}{n_\leaf} \sum_{i : X_i \in \leaf}
      \lpimlog(X_i,Y_i,\hat{p}).
\end{equation*}
We drop the superscript $\leaf$ when it is clear from context.

The PIM-entropy has a closed form, given by the following lemma and proved in Appendix~\ref{app:proofpentropy}.
Like the entropy, it can therefore be evaluated from the four counts of the confusion matrix in constant time, so
\pimcart has the same computational cost as \cart.
\begin{restatable}{lemma}{lemPimEntropy}\label{lem:pim-entropy}
  Let $\lambda > 0$ and let $\leaf$ be a leaf containing at least one data point. The PIM-entropy has the closed form
  \begin{equation*}
      \pimentropy(\confusionmatrix) =
      \begin{cases}
        H(m)  & \text{if } \Lambda \leq 0 , \\[4pt]
        u(s_{\star}) + v(t_{\star}) + \lambda    & \text{if } \Lambda > 0 ,
      \end{cases}
  \end{equation*}
where
\begin{align*}
  & \Lambda = p_{0,0}p_{1,1}-p_{0,1}p_{1,0} - \lambda(p_{0,0}+p_{0,1})(p_{1,0}+p_{1,1}),
  \\
  & u(s) = -p_{1,1}\log(1-s) - p_{0,1}\log s - \lambda s,
  \\
  & v(t) = -p_{0,0}\log(1-t) - p_{1,0}\log t - \lambda t,
\end{align*}
and $s_\star$ and $t_\star$ are, respectively, the larger roots of the following quadratics:
\begin{align*}
    &\varphi(s) = \lambda s^2 + (p_{0,1}+p_{1,1}-\lambda)\,s - p_{0,1},
    \\
    &\psi(t)    = \lambda t^2 + (p_{1,0}+p_{0,0}-\lambda)\,t - p_{1,0}.
\end{align*}
\end{restatable}
As for \cart, \pimcart then chooses each split to minimize the average PIM-entropy of the two children, weighted by
their numbers of points.

\section{Bound on Conversion Accuracy}
\label{sec:conversion_bound}

The \convert convert algorithm does not take the option to defer into
account when choosing the tree structure. Intuitively, this should be
fine if the accuracy of the black box is sufficiently homogeneous over
the domain: we can defer any region on which the tree ends up having low
accuracy without concern for whether this is a good or bad region for
the black box. In this section we formalize this intuition for
converting based on the Lagrangian problem.

The result will actually apply, not just to decision trees, but to any
model class $\simpleF = \{(\partition,\sigma) \mid \partition \in
\partitionclass, \sigma : \partition \to \{0,1,\bot\}\}$ with models
that can be described by a finite partition $\partition =
\{\cell_1,\ldots,\cell_{|\partition|}\}$ of $\domain$ from a class of
candidate partitions $\partitionclass$, together with an extended
labeling function $\sigma$ chosen from all such labeling functions.

We measure the inhomogeneity of the black box with respect to the
possible partitions in $\partitionclass$ by:
\begin{equation}\label{eqn:inhomogeneity_def}
  \begin{split}
  \alpha = \sup_{\partition \in \partitionclass}
  \sum_{\cell \in
  \partition} \Big(&P(X \in \cell, b(X) \neq Y)\\
  &- P(X \in \cell)P(b(X) \neq Y)\Big)_+,
  \end{split}
\end{equation}
where $z_+ = \max\{z,0\}$. Each summand is the excess error of $b$ in a
region $\cell$ over the error $b$ would make if its mistakes were
independent of $\cell$.

We will assume that a learning algorithm (e.g.\ \cart) selects a
partition $\hat \partition \in \partitionclass$ (e.g.\ as induced by a
tree) that is close to minimizing the risk if we never defer:
\begin{equation}\label{eqn:cart_opt_error}
  \underbrace{\min_{\sigma\, :\, \hat \partition \to \{0,1\}}
  R\big((\hat \partition,\sigma)\big)}_{R(\hat \partition)}
  = \underbrace{\inf_{\substack{\partition \in \partitionclass,\\\sigma\, :\, \partition
  \to \{0,1\}}}
R\big((\partition,\sigma)\big)}_{R^*}
    ~+~ \epsilon,
\end{equation}
for some small $\epsilon$.
In addition, suppose that, for all $\lambda \geq 0$, we have a
conversion strategy $\hat \sigma_\lambda$ that approximately minimizes
the population version of the Lagrangian conversion problem:
\begin{equation}\label{eqn:conversion_with_slack}
  \Phi_\lambda\big((\hat \partition,\hat \sigma_\lambda)\big)
  \leq
  \min_\sigma \Phi_\lambda\big((\hat \partition,\sigma)\big)
  + \delta,
\end{equation}
where $\Phi_\lambda(f) = R(f_{|b}) + \lambda U(f)$ denotes the
population version of the Lagrangian objective for any $f \in \simpleF$,
and $\delta$ is small. For instance, we have in mind that $\hat
\sigma_\lambda$ solves the corresponding empirical conversion
problem~\eqref{eqn:CARTLagrangianConvert}.

\begin{figure}
  \centering
  \definecolor{cpblue}{HTML}{0072B2}
\definecolor{hullgreen}{HTML}{009E73}
\definecolor{gapred}{HTML}{D55E00}
\begin{tikzpicture}[x=4.3cm, y=14cm, >=stealth,
    lbl/.style={font=\footnotesize},
    gamma/.style={cpblue, line width=1.2pt},
    gammastar/.style={hullgreen, line width=1.2pt, dash pattern=on 5pt off 2.5pt},
    support/.style={black!60, thin},
    guide/.style={black!50, densely dotted, thin},
    leader/.style={black!35, very thin}]
  \draw[->, black!70] (0,0) -- (1.08,0) node[right, lbl, black] {$U(f)$};
  \draw[->, black!70] (0,0) -- (0,0.24) node[above, lbl, black] {$R(f_{|b})$};
  \draw[support] (0,0.17) -- (0.85,0)    %
    node[pos=0.18, below=0.5pt, lbl, black] {$s_1$};
  \draw[support] (0,0.10) -- (0.5,0)     %
    node[pos=0.25, below=0.5pt, lbl, black] {$s_2$};
  \draw[black!70] (-0.012,0.17) -- (0,0.17);
  \draw[black!70] (-0.012,0.10) -- (0,0.10);
  \node[left, lbl] (philab) at (-0.075,0.17) {$\Phi_\lambda\big((\hat \partition,\hat \sigma_\lambda)\big)$};
  \node[left, lbl] (phistarlab) at (-0.075,0.10) {$\Phi^*_\lambda$};
  \draw[leader] ([xshift=-1.4pt]philab.east) -- (-0.05,0.17);
  \draw[leader] (-0.03,0.17) -- (-0.012,0.17);
  \draw[leader] ([xshift=-1.4pt]phistarlab.east) -- (-0.012,0.10);
  \draw[guide] (0,0.2) -- (-0.06,0.2);
  \draw[guide] (0,0.15) -- (-0.06,0.15);
  \draw[<->, gapred, thick] (-0.04,0.15) -- (-0.04,0.2);
  \node[above, lbl, gapred] at (-0.04,0.2) {$\epsilon$};
  \draw[guide] (0.45,0) -- (0.45,0.0214);
  \draw[guide] (0.25,0) -- (0.25,0.05);
  \draw[black!70] (0.45,0) -- (0.45,-0.005);
  \draw[black!70] (0.25,0) -- (0.25,-0.005);
  \node[lbl, anchor=base] at (0.45,-0.023) {$t$};
  \node[lbl, anchor=base] at (0.25,-0.023) {$t'$};
  \draw[gammastar] (0,0.15) -- (0.1,0.1) -- (0.25,0.05) -- (0.6,0) -- (1,0);
  \draw[gamma] (0,0.2) -- (0.15,0.15) -- (0.35,0.1) -- (0.6,0.05) -- (1,0);
  \draw[<->, gapred, thick] (0.45,0.0214) -- (0.45,0.08);
  \draw[gamma] (0.62,0.215) -- (0.70,0.215) node[right, lbl, black] {$\gamma(t)$};
  \draw[gammastar] (0.62,0.19) -- (0.70,0.19) node[right, lbl, black] {$\gamma^*(t)$};
\end{tikzpicture}
  \caption{Example showing the lower convex hulls of achievable
    trade-offs between uncoverage and risk for the conversion procedure
    ($\gamma$) and the oracle ($\gamma^*$). In this case, the black box
    is perfect, so $\alpha = 0$. We also take zero conversion error:
    $\delta = 0$, which implies that $\epsilon = \gamma(0) -
    \gamma^*(0)$. The line $s_1 = \{r + \lambda u = \Phi_\lambda((\hat
    \partition,\hat \sigma_\lambda)) \mid (u,r) \in \reals^2\}$ supports
    $\gamma$ at $t$, and the line $s_2 = \{r + \lambda u = \Phi^*_\lambda \mid
    (u,r) \in \reals^2\}$ supports $\gamma^*$ at~$t'$. See
    Appendix~\ref{app:details_conversion_gap_figure} for further details.}
  \label{fig:conversion_gap}
\end{figure}
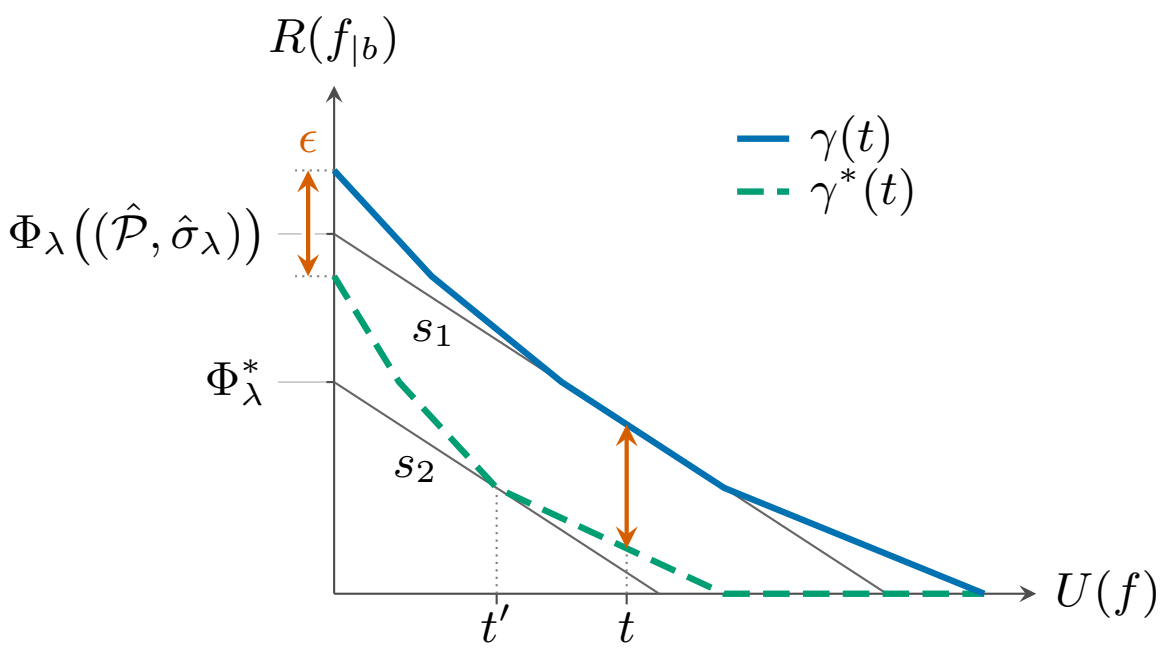

We will consider the lower convex hull of trade-offs between uncoverage
$U(f)$ and risk $R(f_{|b})$, which corresponds to the optimal trade-offs
achievable by solving the Lagrangian problem formulation (see
Section~\ref{sec:CPvsLP} and Appendix~\ref{app:lagrangian-suboptimal}).
For the conversion strategy, this applies to $f$ selected from $\{(\hat
\partition, \hat \sigma_\lambda) \mid \lambda \geq 0\}$, and we denote
the lower convex hull by $\gamma(t)$ for $t \in [0,1]$. We compare this
to the lower convex hull $\gamma^*(t)$ of an oracle that optimizes over
all $f \in \simpleF$. Formally, given any model class $\simpleG$ of the
same type as $\simpleF$, the achievable trade-offs between risk and
uncoverage are:
\begin{equation}\label{eqn:possible_U_R_values}
  S(\simpleG)
    = \big\{\big(U(f),R(f_{|b})\big) \mid f \in \simpleG\big\}
    \subset \reals^2.
\end{equation}
Then the lowest risk on the lower convex hull with uncoverage at most $t
\in [0,1]$ is
\begin{equation}\label{eqn:lower_convex_hull}
  \gamma_\simpleG(t)
  =
  \inf\{
    r \mid (u,r) \in \conv(S(\simpleG)), u \leq t
  \}.
\end{equation}
The two lower convex hulls are then defined as:
\begin{align*}
  \gamma(t) &= \gamma_\simpleG(t)
    \qquad \text{for $\simpleG = \{(\hat \partition,
            \hat \sigma_\lambda) \mid \lambda \geq 0\}$},\\
  \gamma^*(t) &= \gamma_\simpleG(t)
  \qquad \text{for $\simpleG = \simpleF$.}
\end{align*}
Figure~\ref{fig:conversion_gap} illustrates $\gamma$ and $\gamma^*$ for an
example.

\begin{theorem}\label{thm:one_eight_bound}
  Suppose $\alpha,\epsilon$ and $\delta$ satisfy
  \eqref{eqn:inhomogeneity_def}, \eqref{eqn:cart_opt_error} and
  \eqref{eqn:conversion_with_slack}. Then the suboptimality of the
  conversion strategy at any maximum uncoverage $t \in [0,1]$ is at
  most
  \[
    \gamma(t) - \gamma^*(t)
    \leq R(\hat \partition)(1-2R^*) + \delta + \alpha
    \leq \frac{1}{8} (1 + 2 \epsilon)^2 + \delta + \alpha.
  \]
\end{theorem}

This result is easiest to interpret when we think of $\epsilon,\delta$
and $\alpha$ as being $0$. Then $R(\hat \partition) = R^*$ and the bound
shows that the suboptimality of the converting algorithm, compared to
the oracle, is at most $1/8 = 12.5\%$, or even smaller if $R^*$ is far
from $1/4$. Although the conversion approach may seem simplistic, this
means there is limited room to improve on it, because no other
learning algorithm (over the same model class) can beat the oracle.

The case that $\alpha = 0$ may in fact hold exactly, for instance if the
black box is perfect, i.e.\ when $b(X) = Y$ almost surely. But if the
performance of the black box is not homogeneous across the domain, then
$\alpha$ can also be large and then the bound will be vacuous.

The bound is a deterministic statement that holds for any distribution
$P$ for $(X,Y)$. This includes in particular the case that $P$ is equal
to the empirical distribution $P_n$, i.e.\ the uniform distribution over
the sample~$S$. In this case the statements apply to the empirical risk
and the empirical uncoverage. If $\hat \sigma_\lambda$ indeed solves
\eqref{eqn:CARTLagrangianConvert}, we then get $\delta = 0$, and if
$\hat \partition$ is close to minimizing the empirical risk we also have
$\epsilon \approx 0$.

\subsection{Proof of Theorem~\ref{thm:one_eight_bound}}

The proof breaks down into two lemmas, whose formal proofs are deferred to
Appendix~\ref{app:conversion_bound_proof_details}.

Let $\Phi^*_\lambda = \inf_{f \in \simpleF} \Phi_\lambda(f)$. Then the
first lemma is a convex duality result:
\begin{restatable}{lemma}{lemReduction}\label{lem:reduction}
  For any $t \in [0,1]$, there exists $\lambda \geq 0$ (with $-\lambda$
  a subgradient of $\gamma$ at~$t$) such that
  \[
    \gamma(t) - \gamma^*(t)
      \leq
      \Phi_\lambda\big((\hat \partition,\hat \sigma_\lambda)\big)
      - \Phi^*_\lambda.
  \]
\end{restatable}
\emph{Proof sketch.} This is illustrated in
Figure~\ref{fig:conversion_gap}: $s_1$ supports $\gamma$ at $t$ with
slope $-\lambda$, and the parallel line $s_2$ that supports $\gamma^*$
at some $t'$ is a lower bound on $\gamma^*$ at~$t$. The vertical
distance between the two lines therefore bounds $\gamma(t) -
\gamma^*(t)$.

The second lemma continues from the first:
\begin{restatable}{lemma}{lemDualGap}\label{lem:dual_gap}
  For any $\lambda \geq 0$,
  \[
    \Phi_\lambda\big((\hat \partition,\hat \sigma_\lambda)\big)
      - \Phi^*_\lambda
    \leq R(\hat \partition)(1-2R^*) + \delta + \alpha.
  \]
\end{restatable}
\emph{Proof sketch.} Consider the case that $\alpha$ and
$\delta$ are both $0$, so the black box is equally good on every region
$\cell$. Then optimal conversion at any $\lambda$ is at least as good as
$\min\{R(\hat \partition), R(b) + \lambda\}$, by either never or always
deferring. Conversely, the oracle can gain at most a factor $(R(b) +
\lambda)/(1/2)$ compared to $R^*$ by deferring in any region $\cell$
(assuming this factor is less than $1$),
because even without deferring the majority vote already achieves error
at most $1/2$. The difference gives a bound of $\min\{R(\hat
\partition), R(b) + \lambda\} - R^*\min\{1,(R(b) + \lambda)/(1/2)\}$.
Taking the worst case over $R(b) + \lambda$ then completes the proof.

Finally, combining Lemmas~\ref{lem:reduction} and \ref{lem:dual_gap}
gives the first inequality of Theorem~\ref{thm:one_eight_bound}.
Plugging in \eqref{eqn:cart_opt_error} and using that
$R^*=1/4-\epsilon/2$ maximizes the bound then gives its second
inequality.

\section{Experiments}
In this section, we discuss our experimental results. As is common, we
present our results in terms of accuracy and coverage (rather than risk
and uncoverage).

Datasets and protocols are detailed in Appendix~\ref{app:datasets}; Appendices~\ref{app:exp:convert},
\ref{app:exp:frost} and~\ref{app:pimloss} cover the cost of the Lagrangian relaxation, a comparison with decision lists,
and an alternative scoring criterion.

\label{sec:experiments}

\subsection{Comparison of the Three Methods}
\label{sec:exp:trackb}
We compare the
three methods from Section~\ref{sec:decision-tree}. \convert uses the
regular \cart algorithm with either the
Gini index or entropy as a splitting criterion, \pimcart uses $ \pimentropy $ instead. Finally \branches performs exact
search. This gives
four procedures in total.

\paragraph{Setup}
Our benchmark consists of $20$ datasets drawn from the benchmarks of \citet{Chaou_R_B25BranchesEfficientlyseeking} and \citet{Frost_L_M_M24Partiallyinterpretablemodels} (Table~\ref{tab:datasets}). \branches\ requires binary features, so
continuous attributes are thresholded at their median and categorical ones are dummy-encoded. Every method is refit at each of $20$
values of $\lambda \in [0, 1/2]$ with a maximum depth of $10$ and the same complexity penalty
$\gamma = 0.01$, applied as the search penalty for \branches\ and as \texttt{ccp\_alpha} for
\cart.
\branches\ terminates for all values of $\lambda$ on $18$ of the $20$ datasets. The exceptions are \texttt{mushroom} and \texttt{ficobin}, the two largest datasets, on which the search exhausts the available memory (Appendix~\ref{app:datasets}). \citet{Chaou_R_B25BranchesEfficientlyseeking} also report their own solver reaching timeout on \texttt{mushroom} under binary encoding. We also set aside the three \textsc{monk} problems, on which greedy splitting is known to fail (see below), so our averages are over the remaining $15$ datasets.

Each method produces multiple hybrid trees sitting at different accuracy/coverage trade-offs. We score them in the
following manner. For each value of the coverage, we select the best model obtaining at least this value of coverage on
the testing data.
Integrating this over $ [0,1] $ gives us our aggregate score. We also report another criterion, the best accuracy obtained for coverage set at
$t= \frac{1}{4}, \frac{1}{2}, \frac{3}{4} $ by selecting the best model among those with training coverage $ \geq t $,
for which we report both the
test accuracy and test coverage. On about half of the datasets, the methods are within a fraction of a percent of error. The notable exceptions are
the \textsc{monk} datasets in which \branches outperforms other methods. This is expected because they are parity-like problems on
which greedy splitting is known to fail because individual features
are not informative about the label by themselves.

\begin{figure}[t]
  \centering
  \includegraphics[width=\linewidth]{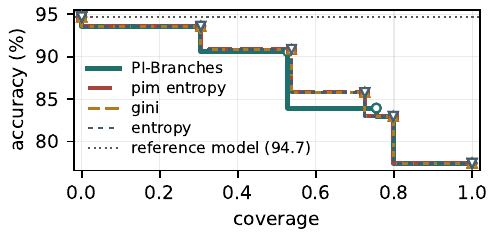}
  \caption{Coverage/accuracy on the held-out split of \texttt{spambase}. Markers are the models obtained for the
      different values of $\lambda$.}
  \label{fig:trackb-spambase}
\end{figure}

\paragraph{One dataset in full}
Figure~\ref{fig:trackb-spambase} shows our methods on the \texttt{spambase} dataset. The four
frontiers coincide almost everywhere; the numbers behind it, and the same breakdown for every
other dataset, are in Table~\ref{tab:trackb-full-A}.

\paragraph{Across datasets}
Table~\ref{tab:trackb-summary} averages the same quantities over the $15$ datasets that remain once the three \textsc{monk} problems and the two datasets on which \branches does not terminate are set aside, and the picture is the same there: the four
procedures lie within $0.60\%$ error of one another on the aggregate
score and within $1.12\%$ on the tightest coverage threshold. Per
dataset, the methods are within $1.2\%$ error of one another on $10$ of
the $15$ datasets; the larger gaps (\texttt{sonar}, \texttt{ionosphere},
\texttt{heart\_statlog}, \texttt{tictactoe}, \texttt{hepatitis}) arise
on small datasets and vary in which method they favor. Complete results by dataset can be found in Appendix~\ref{app:trackb-tables}.

\begin{table}[t]
\centering\small
\setlength{\tabcolsep}{3pt}
\begin{tabular}{lrrrr}
\toprule
 & \branches & \texttt{gini} & \texttt{entropy} & \pimcart \\
\midrule
aggregate score & 85.39 & 84.84 & 84.86 & \textbf{85.44} \\
$c \geq 0.25$ & \textbf{86.08} & 85.35 & 85.46 & 85.15 \\
$c \geq 0.50$ & \textbf{85.36} & 85.14 & 85.25 & 84.62 \\
$c \geq 0.75$ & \textbf{84.13} & 83.93 & 83.93 & 83.01 \\
\bottomrule
\end{tabular}

\caption{Comparison between methods, averaged over the 15 datasets on which every method completed, \emph{excluding} the
three \textsc{monk} problems. Per dataset results are in Appendix~\ref{app:trackb-tables}.}
\label{tab:trackb-summary}
\end{table}

In particular, \pimcart and \branches, which take the black box into
account while growing the tree, show no advantage over \convert on real
data. That is a property of the data rather than of the methods. In
Appendix~\ref{app:exp:synthetic}, we construct a dataset on which the
accuracy of the black box is deliberately inhomogeneous across the
domain, and there \pimcart and \branches do outperform \convert.

\subsection{Application to Concept Bottleneck Models}
\label{sec:exp:cbm}

Concept bottleneck models (CBMs)
\citep{Koh_N_T_M_P_K_L20Conceptbottleneckmodels} predict in two stages:
a (potentially complicated) map from raw features to a layer of
human-understandable concepts, followed by an interpretable model $f$
that maps concepts to predictions. This allows CBMs to deal with complex
features such as pixels, to support intervention on concepts and to be
human-interpreted through their concepts. Concepts are either
\emph{hard} (binary) or \emph{soft} (continuous scores).

CBMs tend to have worse accuracy compared to an unconstrained black box.
And, in spite of the promise of interpretability, the concept-to-label
part is often a neural network, which is usually not considered
inherently interpretable. We replace this second stage by a partially
interpretable tree on the concepts, which defers to an unconstrained
black box. This allows us to trade off coverage for accuracy, and match
the accuracy of existing CBMs. When the hybrid model does not defer, the
path is a sufficient explanation from the concepts, as discussed in the
introduction. More precisely, two data points that follow the same path
will get the same prediction from the tree. If that prediction is to not
defer, they will get the same prediction from the hybrid model.

\begin{figure}[t]
  \centering
  \includegraphics[width=\linewidth]{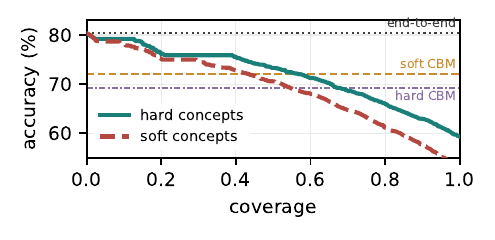}
  \caption{CUB-200, 200-way species. We represent the different accuracy-coverage tradeoffs obtained by our hybrid
  models.}
  \label{fig:cub-bottleneck}
\end{figure}

\begin{figure}[t]
  \centering
  \includegraphics[width=0.85\linewidth]{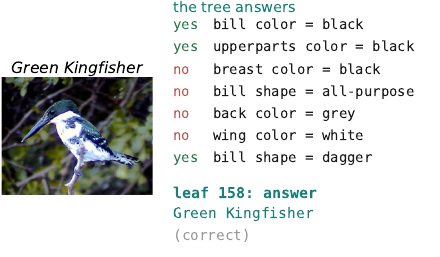}
  \caption{The output of the hybrid model on one image. The tree does
    not defer, so this path is the whole explanation: any image whose
concepts follow the same path gets the same prediction.}
\label{fig:bird_explanation}
\end{figure}

We demonstrate this approach on the CUB-200 bird dataset
\citep{Wah_B_W_P_B25CaltechUCSDbirds2002011}: $200$ species to be
identified from $112$ human-interpretable concepts, in the setting of
\citet{Havas_P_D22Addressingleakageconcept}
(Appendix~\ref{app:datasets}). The baseline CBM maps the predicted
concepts to the classes with a two-layer network of approximately $60$k
parameters. Our
hybrid model is a decision tree on those same $112$ concepts, deferring
to an unconstrained Inception~v3 model on pixels. Since a depth-$d$ tree
has at most
$2^d$ leaves, we consider a depth of $8$ so there is at least
one leaf per species. The \convert method then gives us a family of
partially interpretable decision trees. The possible trade-offs between
coverage and accuracy are shown in
Figure~\ref{fig:cub-bottleneck}. With hard concepts, the tree matches
the accuracy of the soft CBM at $57\%$ coverage, which means that while
not losing any accuracy, we can provide a full explanation for well over
half of the data. Figure~\ref{fig:bird_explanation} shows an image on
which the tree predicts and Appendix~\ref{app:bird-deferred} an image on
which the tree defers.

\section{Conclusion}
\label{sec:conclusion}

We found all three proposed algorithms to have similar performance in
experiments on real-world data. \convert has the advantage that it works
with off-the-shelf \cart software. At the extra effort of modifying
standard \cart, \pimcart has the same computational cost and can, in
principle, have an advantage if the accuracy of the black box is
inhomogeneous across the domain. \branches is the method of choice on
datasets where greedy splitting fails (such as parity-like problems),
but it may become computationally infeasible on larger datasets.

Applied to concept bottleneck models, we showed that partially
interpretable trees can provide an alternative to two-layer
neural networks with a large number of parameters.

\paragraph{Limitations and future work}

Our bound and the PIM-entropy criterion are stated for binary
classification, and the bound becomes vacuous when the black box is
strongly inhomogeneous. Natural next steps are to estimate the
inhomogeneity from data to decide when deferral-aware splitting is
worthwhile and extend our criterion and analysis to multi-class
problems. Finally, the application of partially interpretable trees in
concept bottleneck models looks very promising, and deserves a
systematic study across more benchmarks.

\FloatBarrier
\subsubsection*{Acknowledgements}

Schwartz, Berton and Van Erven were all supported by the Netherlands
Organization for Scientific Research (NWO) under grant ID
\url{https://doi.org/10.61686/OWREN85146}.

\subsection*{AI Use Statement}

We used Claude Code with various versions of the Opus and Fable models
between February and October 2026.

The algorithms in Section~\ref{sec:decision-tree} were designed without
the use of AI.

Theorem~\ref{thm:one_eight_bound} was derived with the help of AI.
Lemma~\ref{lem:pim-entropy} was proved entirely without AI assistance.
Their write-ups in the paper (statements and proofs) were written by
hand.

For the experiments, the methods were implemented by AI, which was also
used to provide feedback on research methodology and experimental setup, and to write the data-preprocessing code, which
applies only the preprocessing described in Appendix~\ref{app:datasets}. We
checked the AI-written code in several ways. The authors reviewed the data-preprocessing pipeline, the implementation of
every method and the code of every
experiment as well as the diff between our implementation of \branches and the original \origbranches solver. Each
result in the paper is produced in two stages: one script computes and stores the results, and another generates the
table or figure from the stored results. Each script has a limited scope and was reviewed individually.

The paper was mostly written without AI, but some parts were initially
drafted by AI and then rewritten:
Sections~\ref{sec:intro}--\ref{sec:conversion_bound} were written
without AI. Section~\ref{sec:experiments} was drafted with AI, but then
fully rewritten by hand. Appendix~\ref{app:lagrangian-suboptimal} was
written by hand; Appendix~\ref{app:lagrangian-family} was initially
drafted by AI, but then rewritten. Appendix~\ref{app:Branches} was
drafted by AI and then rewritten. Appendices~\ref{app:knapsack}
and~\ref{app:proofpentropy} were written without AI.
Appendix~\ref{app:details_conversion_gap_figure} was written by hand. The remaining
parts of Appendix~\ref{app:sec:exp} were drafted by AI and then
rewritten.

The figures in the paper were all produced with AI, and AI was used for
proofreading. AI was also used to identify additional related
literature. Every suggested reference was read and verified by the
authors, and the bibliography was compiled manually.

We have carefully reviewed all AI-assisted work. The conclusions
presented here are our own. We take responsibility for the final content
of this work, including text, claims, or artifacts produced with the aid
of AI.

\bibliographystyle{plainnat}
\bibliography{ref}

\clearpage
\appendix
\thispagestyle{empty}
\onecolumn
\aistatstitle{Partially Interpretable Decision Trees: \\
Supplementary Materials}

\section{Supplement to Section~\ref{sec:pim}}
\label{app:sec:pim}
\subsection{Potential Suboptimality of Lagrangian Optimization}
\label{app:lagrangian-suboptimal}

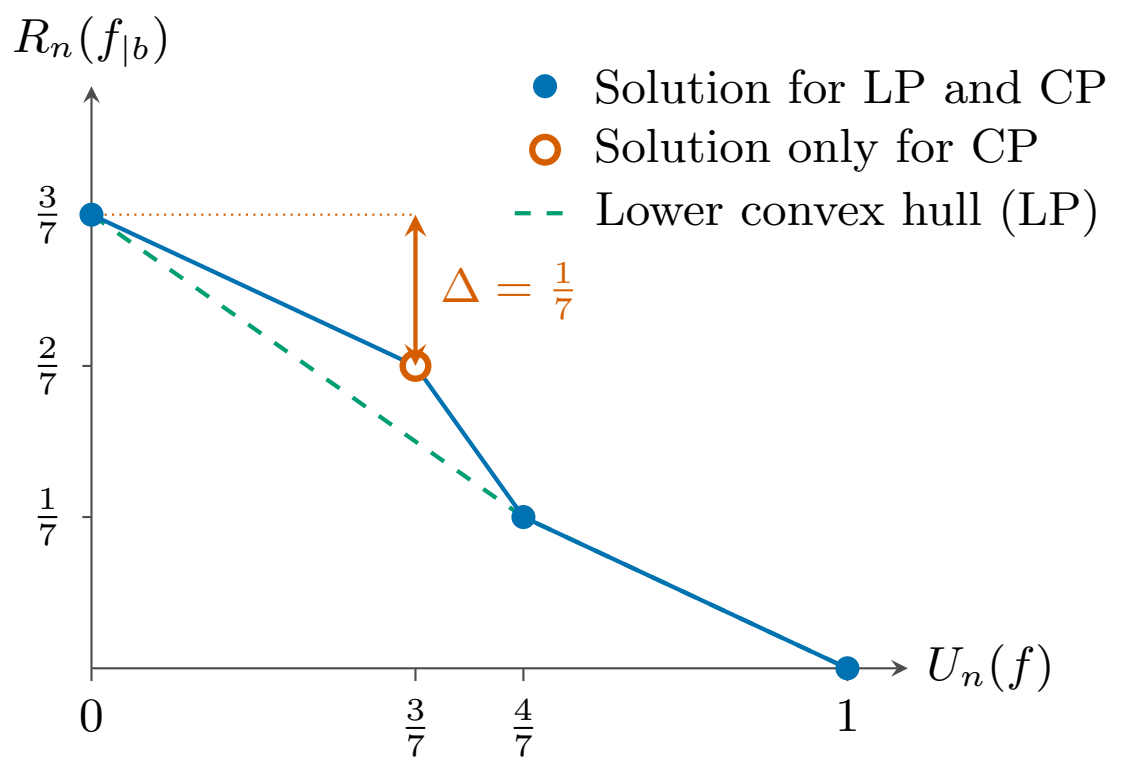
\begin{figure}[!hb]
  \centering
  \definecolor{cpblue}{HTML}{0072B2}
\definecolor{hullgreen}{HTML}{009E73}
\definecolor{gapred}{HTML}{D55E00}
\begin{tikzpicture}[x=5cm, y=7cm, >=stealth,
    dot/.style={circle, fill=cpblue, inner sep=1.6pt},
    miss/.style={circle, draw=gapred, very thick, fill=white, inner sep=1.7pt},
    lbl/.style={font=\footnotesize}]
  \draw[->, black!70] (0,0) -- (1.08,0) node[right, lbl, black] {$U_n(f)$};
  \draw[->, black!70] (0,0) -- (0,0.55) node[above, lbl, black] {$R_n(f_{|b})$};
  \foreach \u/\l in {0/0, 0.4286/{\tfrac37}, 0.5714/{\tfrac47}, 1/1} {
    \draw[black!70] (\u,0) -- (\u,-0.012) node[below, lbl, black] {$\l$};
  }
  \foreach \r/\l in {0.1429/{\tfrac17}, 0.2857/{\tfrac27}, 0.4286/{\tfrac37}} {
    \draw[black!70] (0,\r) -- (-0.012,\r) node[left, lbl, black] {$\l$};
  }
  \draw[dashed, thick, hullgreen] (0,0.4286) -- (0.5714,0.1429) -- (1,0);
  \draw[thick, cpblue] (0,0.4286) -- (0.4286,0.2857) -- (0.5714,0.1429) -- (1,0);
  \node[dot] (A) at (0,0.4286) {};
  \node[dot] (C) at (0.5714,0.1429) {};
  \node[dot] (D) at (1,0) {};
  \node[miss] (B) at (0.4286,0.2857) {};
  \draw[densely dotted, gapred] (A) -- (0.4286,0.4286);
  \draw[<->, gapred, thick] (0.4286,0.2857) -- (0.4286,0.4286)
    node[midway, right=1pt, lbl, gapred] {$\Delta = \tfrac17$};
  \node[dot] at (0.6,0.55) {};
  \node[right, lbl] at (0.64,0.55) {Solution for LP and CP};
  \node[miss] at (0.6,0.49) {};
  \node[right, lbl] at (0.64,0.49) {Solution only for CP};
  \draw[dashed, thick, hullgreen] (0.56,0.43) -- (0.64,0.43)
    node[right, lbl, black] {Lower convex hull (LP)};
\end{tikzpicture}
  \caption{Empirical risk
    $R_n(f_{|b})$ against empirical uncoverage $U_n(f)$ for all Pareto
    optimal $f$ in a class $\simpleF$ of decision stumps (depth one
    trees) on a toy dataset of $n = 7$ points with $d = 2$ binary
    features, where $b$ is perfect on deferred points. The orange open
    circle point lies strictly above the lower boundary of the convex
    hull of the frontier, so it is not a solution of the LP for any
    $\lambda$, even though it solves the CP for $t = \tfrac37$. The best
    LP solution with $U_n(f) \leq \tfrac37$ has empirical risk larger by
    $\Delta = \tfrac17 \approx 14\%$.}
    \label{fig:pareto}
\end{figure}

To illustrate which Pareto optimal solutions are not solutions to any
LP, consider the example plot in Figure~\ref{fig:pareto}. It was
obtained by exhaustive search over all datasets of size $n=7$ with two
binary features to maximize the suboptimality $\Delta$. The points
obtainable as the solutions to the LP for some value of $\lambda$ are
the ones that sit on the lower boundary of the convex hull of the Pareto
frontier, so the Pareto-optimal open circle point cannot be a solution to
the LP. To see this, let $\phi_\lambda = \min_{f \in \simpleF}
\,\{R_n(f_{|b}) + \lambda U_n(f)\}$ be the value of the LP at $\lambda$.
Then all solutions to the LP for this $\lambda$ must lie on the line
$\{r + \lambda u = \phi_\lambda \mid r,u \in \reals\}$. But this line must
also be a supporting line of the Pareto frontier: if there was a Pareto
optimal point strictly below the line, then we could move it down by
decreasing $\phi_\lambda$ and find a better solution to the LP. It follows
that all solutions to any LP sit on the lower convex hull of the Pareto
frontier. In the figure, this means that (only) the filled blue points
are obtainable as solutions to the LP for some value of $\lambda$, but
the open circle point is not. If we set a strict upper bound on the
uncoverage, then the resulting suboptimality of the empirical risk for
the best point obtainable as the solution to an LP can be significant:
it is possible to construct a dataset with $n = 4k-1$ points that has a
gap of $\Delta = \frac{k-1}{4k-1} \rightarrow \frac{1}{4}$
(Appendix~\ref{app:lagrangian-family}), and the figure shows its case $k = 2$, which
gives $\Delta \approx 14\%$. For many randomly generated datasets the suboptimality appears to
be much smaller or even non-existent, though. Pragmatically, it is
therefore still reasonable to optimize for the LP when the CP is
computationally infeasible.

\subsection{A Family Whose Gap Approaches $1/4$}
\label{app:lagrangian-family}

Figure~\ref{fig:pareto} shows a dataset on which the gap between the Lagrangian and the
constrained solutions is significant ($\Delta = \tfrac17$). By adding more points to it, we can
widen this gap up to $\tfrac14$. We use two binary features $x_1$ and $x_2$, so a dataset is
fully described by how many points of each label fall into each of the four combinations of
$(x_1, x_2)$. For $k \geq 2$, let the dataset $S_k$ have the following counts:
\begin{center}
\begin{tabular}{ccc}
\toprule
$(x_1, x_2)$ & $y = 0$ & $y = 1$ \\
\midrule
$(0,0)$ & $0$ & $k-1$ \\
$(0,1)$ & $k$ & $0$ \\
$(1,0)$ & $k$ & $0$ \\
$(1,1)$ & $0$ & $k$ \\
\bottomrule
\end{tabular}
\end{center}
So $n = 4k-1$, and $S_2$ is the dataset of Figure~\ref{fig:pareto}. The labels are noiseless:
$y = 1$ exactly when $x_1 = x_2$, which decision stumps of depth $1$ cannot represent.

Let $\simpleF$ be the stumps on $x_1$ or $x_2$, each branch either predicting its majority label
or deferring to $b$, with $b$ perfect on deferred points. Splitting on $x_1$, the branch
$x_1 = 0$ holds $k-1$ points labeled $1$ and $k$ labeled $0$: it has $2k-1$ points, its
majority label is $0$ and it misclassifies $k-1$ points. The branch $x_1 = 1$ holds $k$ of each label:
it has $2k$ points and misclassifies $k$ whichever label it predicts. Splitting on $x_2$ is
symmetric. The four ways to resolve the two branches give, in counts $(nU_n, nR_n)$,
\[
  \underbrace{(0,\; 2k-1)}_{\text{predict both}}, \quad
  \underbrace{(2k-1,\; k)}_{\text{defer } x_1 = 0}, \quad
  \underbrace{(2k,\; k-1)}_{\text{defer } x_1 = 1}, \quad
  \underbrace{(4k-1,\; 0)}_{\text{defer both}},
\]
and these four are exactly the Pareto-optimal points of $\simpleF$ on $S_k$.

\paragraph{The suboptimality gap}
The lower convex hull of the frontier has vertices $(0, 2k-1)$, $(2k, k-1)$ and $(4k-1, 0)$: the
segment between the first two has slope $-\tfrac12$ and passes through height $k - \tfrac12$ at
$nU_n = 2k-1$, so the second Pareto point $(2k-1,\, k)$ lies strictly above it. By the
supporting-line argument of Appendix~\ref{app:lagrangian-suboptimal} it therefore solves no LP,
for any $\lambda$.

Now constrain the uncoverage to $t = \frac{2k-1}{4k-1}$. The constrained problem attains
$R_n = \frac{k}{4k-1}$ at that very point. Every Lagrangian solution is a hull vertex, and the
only one with $U_n \leq t$ is $(0, 2k-1)$, of risk $\frac{2k-1}{4k-1}$. The penalty for
restricting to Lagrangian solutions is thus
\[
  \Delta_k \;=\; \frac{2k-1}{4k-1} - \frac{k}{4k-1} \;=\; \frac{k-1}{4k-1}
  \;\xrightarrow[k \to \infty]{}\; \frac14 .
\]
\FloatBarrier

\section{Supplement to Section~\ref{sec:decision-tree}}
\label{app:sec:tree}
\subsection{The \origbranches and \branches Algorithms}
\label{app:Branches}

We describe the full \origbranches and \branches algorithms. Algorithm~\ref{alg:branches} is \origbranches\ as
described by \citet{Chaou_R_B25BranchesEfficientlyseeking}, with every line that \branches\
changes or adds set \pimdiff{in bold teal}; the other lines are common to both. Running the algorithm
with the highlighted lines removed, or equivalently with $\lambda$ and $b$ absent, recovers
\origbranches\ exactly.

\paragraph{Notation and the dynamic program}
Throughout this section we consider features
$ X = (X^{(1)},\ldots,X^{(q)}) \in \{0,1\}^q $.
A branch $ l $ is a conjunction of clauses on the features of the form
\begin{equation*}
    l = \bigwedge_{v=1}^{S(l)} \ind{X^{(i_v)}=j_v}, \qquad j_v \in \{ 0,1 \},
\end{equation*}
where $ S(l) $ is the number of clauses.
For any data point $ X $, the valuation of $ l $ at $ X $ is defined by
\begin{equation*}
    l (X) = 1 \iff X^{(i_v)} = j_v \ \text{ for all } v \in \{ 1,\ldots,S(l) \}.
\end{equation*}

One should think of a branch as a potential path in a decision tree, and the valuation of a data point in that branch is
1 if that data point would have followed that path. We define the number of points in $ l $ as $ n(l) = \sum_{i=1}^{n}
l(X_i) $. We define the predicted class in $ l $ as
\begin{equation*}
    k^{*}(l) = \argmax_{k\in \{ 0,1 \}}{n_k(l)},
\end{equation*}
where $ n_k(l) = \sum_{i=1}^{n} l(X_i)\ind{Y_i=k} $ is the number of data points in $ l $ that
are of class $ k $. So $ k^{*} $ is the majority class in $ l $, and the fraction of the points
of $ l $ that it misclassifies is the \emph{error rate}
\begin{equation*}
    e(l) = \frac{n(l) - n_{k^{*}}(l)}{n(l)}.
\end{equation*}

To compute an optimal decision tree \origbranches considers a dynamic programming approach. At a branch $ l $ the algorithm
chooses an action $ a $: either the \emph{terminal action} $ \bar{a} $, which stops the branch
and makes it a leaf, or a \emph{split action}, which adds a split and replaces $ l $ by its two
children. The search maintains a state-value $ \mathcal{V}(l) $ for the branch $ l $ and a
state-action value $ \mathcal{Q}(l,a) $ for each action available at it. These are
\emph{estimates}, refined as the search proceeds; we write $ \mathcal{V}^{*}(l) $ and
$ \mathcal{Q}^{*}(l,a) $ for the optimal values they bound.

In particular, we need to be able to evaluate the terminal action. Stopping at $ l $ makes a
mistake on every point of $ l $ not of the majority class, so
\begin{equation*}
    \mathcal{Q}(l,\bar{a}) = \mathcal{Q}^{*}(l, \bar{a})
      = \frac{n(l)}{n}\,e(l) = \frac{n(l) - n_{k^{*}}(l)}{n},
\end{equation*}
the fraction of the whole sample that $ l $ misclassifies. It is exact because stopping leaves
no further decision to make.

For a split action $ a $, the loss decomposes over the two children the split creates, and the
Bellman equation suggests the recursive definition
\begin{equation}\label{eqn:branches-bellman}
    \mathcal{Q}(l,a) = \gamma + \sum_{u} \mathcal{V}(l_u),
    \qquad
    \mathcal{V}(l) = \min_{a} \mathcal{Q}(l,a),
\end{equation}
where $ l_u $ ranges over the two children of $ l $ under $ a $. This determines
$ \mathcal{V}(l) $ only once the children have been evaluated, so the recursion has to be
started somewhere: $ \mathcal{V} $ is initialized by a bound at the branches the search
chooses to expand, and backpropagated through \eqref{eqn:branches-bellman} thereafter.

For the terminal action we have the exact value. For a split action, the best case scenario is
that every point is correctly classified after the split and we only pay the splitting cost
$ \gamma $. This gives the \emph{Purification Bound} of
\citet{Chaou_R_B25BranchesEfficientlyseeking},
\begin{equation}\label{eqn:branches-bound}
    \mathcal{V}(l) = \min\Big\{ \mathcal{Q}(l,\bar{a}),\ \gamma \Big\},
\end{equation}
a lower bound on $ \mathcal{V}^{*}(l) $, the loss of the best subtree rooted at $ l $. When no
split action remains, because $ l $ already uses every feature or has reached the maximum
depth, the bound is exact and
$ \mathcal{V}(l) = \mathcal{V}^{*}(l) = \mathcal{Q}(l,\bar{a}) $.

In particular, when the terminal value satisfies $ \mathcal{Q}(l, \bar{a}) \leq \gamma $, stopping is optimal and the
branch is closed needing no further exploration.

Extending
\origbranches to our framework requires two ingredients. First when evaluating the value of a leaf, one must consider
the possibility of deferring. A deferring leaf makes the black box's mistakes instead of the
tree's, and pays $ \lambda $ for each of its points. Writing
$\pimdiff{n_b(l) = \sum_{i=1}^{n} l(X_i)\ind{b(X_i)=Y_i}}$ for the number of points of $ l $ that the
black box classifies correctly, the leaf takes whichever action is cheaper. That gives the
terminal value of \branches,
\begin{equation}\label{eqn:pibranches-leaf}
    \mathcal{Q}_{\mathrm{pim}}(l,\bar{a}) = \frac{1}{n}\min\Big\{
      \underbrace{n(l) - n_{k^{*}}(l)}_{\text{predict}},\
      \underbrace{n(l) - n_b(l) + \lambda\, n(l)}_{\text{defer}} \Big\},
\end{equation}
which replaces $ \mathcal{Q}(l,\bar{a}) $. Deferring is chosen exactly when
$ \lambda\, n(l) < n_b(l) - n_{k^{*}}(l) $.
Second, the heuristic. The Purification Bound keeps its form,
\begin{equation}\label{eqn:pibranches-bound}
    \mathcal{V}_{\mathrm{pim}}(l) =
      \min\Big\{ \mathcal{Q}_{\mathrm{pim}}(l,\bar{a}),\ \gamma \Big\},
\end{equation}
and remains a valid lower bound (see the paragraph after Algorithm~\ref{alg:branches}).
$ \mathcal{V}_{\mathrm{pim}} $ is backpropagated through
\eqref{eqn:branches-bellman} exactly as $ \mathcal{V} $ is.

The full algorithm is Algorithm~\ref{alg:branches} below. Only the leaf evaluation changes
(lines~\ref{line:nb}--\ref{line:sigma}): with the highlighted lines, $Q$ and $V$ are the
$\mathcal{Q}_{\mathrm{pim}}$ and $\mathcal{V}_{\mathrm{pim}}$ of
\eqref{eqn:pibranches-leaf} and~\eqref{eqn:pibranches-bound}.

\paragraph{Values} Everything below is a \emph{loss}, matching the minimization of
Section~\ref{sec:decision-tree}. A tree $T = (\leaves, \sigma)$ costs
$\sum_{l \in \leaves} \mathcal{Q}(l, \bar a) + \gamma|\leaves|$. All losses are
normalized by the total sample size $n$, not by $n(l)$, so that the contributions of
sibling leaves add. (The implementation maximizes the equivalent reward
$n(l)/n - \mathcal{Q}(l,\bar a)$; the two differ only in sign.) Each branch carries
\begin{itemize}
  \itemsep0pt
  \item $\mathcal{Q}(l, \bar a)$, the exact loss of stopping at $l$;
  \item $\mathcal{V}(l)$, an optimistic \emph{lower} bound on the loss of the best subtree
        rooted at $l$, used to order the search;
  \item $\mathcal{V}_{g}(l)$, the loss of the best subtree actually found so far, an \emph{upper}
        bound, used to pick which sibling to descend into;
  \item a priority queue over attributes, a flag \textsc{terminal} (stopping is provably
        optimal here) and a flag \textsc{complete} (the optimum below $l$ is proved).
\end{itemize}
The search ends when the root is \textsc{complete}; the returned tree is then optimal for the
objective. Capping the depth or the iteration budget forfeits that guarantee, and the
algorithm returns the best tree found instead.

\begin{breakablealgorithm}
\caption{\origbranches, with the \pimdiff{\branches} differences in bold teal}
\label{alg:branches}
\begin{algorithmic}[1]
\Require data $(X_i, Y_i)_{i=1}^{n}$, complexity penalty $\gamma \geq 0$, maximum depth $D$,
         iteration budget $N$
\Require[\pimdiff{PI}] \pimdiff{black box predictions $b(X_1), \dots, b(X_n)$, deferral price
         $\lambda \geq 0$}
\Ensure a tree $T = (\leaves, \sigma)$

\Statex
\Function{EvaluateTerminal}{$l$}
  \State $k^{*}(l) \gets \argmax_{k} n_k(l)$
  \State \pimdiff{$n_b(l) \gets \#\{i : l(X_i) = 1,\ b(X_i) = Y_i\}$}
         \label{line:nb}
  \State $\mathcal{Q}(l, \bar a) \gets \dfrac{1}{n}
         \pimdiff{\min\Big\{} n(l) - n_{k^{*}}(l)
         \pimdiff{,\ n(l) - n_b(l) + \lambda\, n(l) \Big\}}$
         \label{line:Q}
  \State \pimdiff{$\sigma(l) \gets \defer$ \textbf{if} $n(l) - n_b(l) +
         \lambda n(l) < n(l) - n_{k^{*}}(l)$ \textbf{else} $k^{*}(l)$}
         \label{line:sigma}
\EndFunction

\Statex
\Function{Evaluate}{$l$}
  \State \Call{EvaluateTerminal}{$l$}
  \If{$l$ has an unused attribute and $\mathrm{depth}(l) < D$}
    \State $\mathcal{V}(l) \gets \min\big\{\, \mathcal{Q}(l, \bar a),\ \gamma \,\big\}$
           \Comment{Purification Bound}
           \label{line:bound}
    \If{$\mathcal{V}(l) = \mathcal{Q}(l, \bar a)$}
       \State mark $l$ \textsc{terminal} and \textsc{complete}
       \Comment{stopping is provably optimal}
    \EndIf
  \Else
    \State $\mathcal{V}(l) \gets \mathcal{Q}(l, \bar a)$; mark $l$ \textsc{terminal} and \textsc{complete}
  \EndIf
  \State $\mathcal{V}_{g}(l) \gets \mathcal{Q}(l, \bar a)$
\EndFunction

\Statex
\Function{Split}{$l$, attribute $a$}
  \For{$j \in \{0, 1\}$}
    \State $l_j \gets l \wedge \ind{X^{(a)} = j}$, with $n(l_j), n_1(l_j)$
           \pimdiff{and $n_b(l_j)$} obtained from the parent's counts
    \State \Call{Evaluate}{$l_j$}
  \EndFor
  \State push $a$ on $l$'s queue with key $\mathcal{V}(l_0) + \mathcal{V}(l_1)$
         \Comment{smallest key first}
\EndFunction

\Statex
\Function{Select}{\,}
  \State $l \gets$ root; $\mathit{path} \gets [\,l\,]$
  \While{$l$ has children \textbf{and} $l$ is not \textsc{complete}}
    \State $a \gets$ head of $l$'s queue
    \State $l \gets$ the child of $l$ under $a$ with the largest $V_{g}$
           \Comment{the least settled sibling}
    \State append $l$ to $\mathit{path}$
  \EndWhile
  \State \Return $l, \mathit{path}$
\EndFunction

\Statex
\Function{Expand}{$l$}
  \For{each attribute $a$ unused on $l$}
    \State \Call{Split}{$l$, $a$}
  \EndFor
  \State $\mathcal{V}(l) \gets \min\big\{\, \mathcal{Q}(l, \bar a),\
         \gamma + \min_{a} \big( \mathcal{V}(l_0^{a}) + \mathcal{V}(l_1^{a}) \big) \,\big\}$
  \If{$\mathcal{V}(l) = \mathcal{Q}(l, \bar a)$} \State mark $l$ \textsc{terminal} and \textsc{complete}
  \ElsIf{every child under the best attribute is \textsc{complete}}
         \State mark $l$ \textsc{complete}
  \EndIf
\EndFunction

\Statex
\Function{Backpropagate}{$\mathit{path}$}
  \For{$l$ from the parent of the selected branch up to the root}
    \State discard the \textsc{complete} children from $l$'s queue, folding their values
           into the key
    \State $a^{*}(l) \gets \argmin_{a} \big(\gamma + \mathcal{V}_{g}(l_0^{a}) +
           \mathcal{V}_{g}(l_1^{a})\big)$;
           \ $\mathcal{V}_{g}(l) \gets \min\big\{ \mathcal{Q}(l, \bar a),\ \gamma +
           \mathcal{V}_{g}(l_0^{a^{*}}) + \mathcal{V}_{g}(l_1^{a^{*}})\big\}$
    \State $\mathcal{V}(l) \gets \min\big\{ \mathcal{Q}(l, \bar a),\ \gamma + \min_{a} \big(
           \mathcal{V}(l_0^{a}) + \mathcal{V}(l_1^{a}) \big) \big\}$
    \If{$\mathcal{V}(l) = \mathcal{Q}(l, \bar a)$} \State mark $l$ \textsc{terminal} and \textsc{complete}
    \ElsIf{every child under the best attribute is \textsc{complete}}
           \State mark $l$ \textsc{complete}
    \EndIf
  \EndFor
\EndFunction

\Statex
\Statex \textbf{Main loop}
\State $\mathrm{root} \gets$ the empty conjunction; \Call{Evaluate}{$\mathrm{root}$};
       \Call{Expand}{$\mathrm{root}$}
\For{$i = 1$ \textbf{to} $N$}
  \If{root is \textsc{complete}} \State \textbf{break}
         \Comment{the optimum is proved}
  \EndIf
  \State $l, \mathit{path} \gets$ \Call{Select}{\,}
  \State $\mathit{sib} \gets$ the sibling of $l$ under its parent's best attribute
  \If{$l$ is not \textsc{complete}} \State \Call{Expand}{$l$} \EndIf
  \If{$\mathit{sib}$ is neither \textsc{complete} nor expanded}
      \State \Call{Expand}{$\mathit{sib}$}
  \EndIf
  \State \Call{Backpropagate}{$\mathit{path}$}
\EndFor
\State $\leaves \gets$ the branches reached by following $a^{*}$ from the root to a
       \textsc{terminal} branch
\State \Return $T = (\leaves, \sigma)$
       \Comment{\pimdiff{$\sigma(l) = \defer$ on the deferring leaves}}
\end{algorithmic}
\end{breakablealgorithm}

\paragraph{Why the optimistic bound is still valid under deferral}
Line~\ref{line:bound} bounds the loss of \emph{any} subtree rooted at $l$ from below by
$\gamma$: a subtree with more than one leaf pays for at least one split, and in the best case
makes no mistakes at all. Deferring cannot lower that floor, because by line~\ref{line:Q} the
loss of a leaf is non-negative whichever action it takes: the deferring alternative is
$\big(n(l') - n_b(l') + \lambda n(l')\big)/n \geq 0$ for every
$l' \subseteq l$, a count of mistakes plus a non-negative price, and it is zero only
when the black box is perfect on $l'$ and $\lambda = 0$. The bound is therefore
admissible for \branches\ as it stands, which is why the search remains exact and why no line
of \textsc{Evaluate} outside the leaf value had to change.

\paragraph{Cost} The lattice of branches is shared: two paths with the same set of clauses in
a different order are the same node, which is what makes the search tractable at all. Memory
is nonetheless the binding constraint in our experiments rather than time, and it grows with
the number of binary attributes. On the two largest datasets of Section~\ref{sec:exp:trackb}, \texttt{mushroom} and \texttt{ficobin}, the search exhausted the available memory before proving optimality; we report those datasets as
incomplete, there is no optimality guarantee on those.
\FloatBarrier
\subsection{The Knapsack Formulation of \convert}
\label{app:knapsack}

Section~\ref{sec:cart-convert} reduces the constrained conversion of a fixed tree to a $0/1$
knapsack whose items are the leaves, a leaf being included exactly when the tree defers on it.
The weights and values are as follows.

The weight of each
leaf~$\leaf$ is $w_\leaf = \frac{n_\leaf}{n}$, where $n_\leaf =
|\{i:X_i \in \leaf\}|$ is the number of data points in $\leaf$, so the
sum of the weights for leaves that are included in the knapsack should
not exceed $t$.\footnote{Without loss of generality, $t$ may be rounded
  down to the closest multiple of $1/n$, so it effectively has $n+1$
possible values.} The value of each leaf is $v_\leaf = \frac{1}{n}
\min_{\hat y \in \{0,1\}} \sum_{i:X_i \in \leaf}
\big(\ell_{0/1}(Y_i,\hat y) - \ell_{0/1}(Y_i,b(X_i))\big)$, which is the
reduction in empirical risk by deferring for the leaf. Although knapsack
problems are NP-hard in general, in our setting the number of leaves is
typically small enough that the knapsack can be solved exactly by
standard knapsack solvers.

\FloatBarrier
\subsection{The Proof of Lemma~\ref{lem:pim-entropy}}
\label{app:proofpentropy}

\lemPimEntropy*

The condition $\Lambda > 0$ has a simple interpretation. If $0 < m < 1$, then $\Lambda = m(1-m)(J - \lambda)$, where
$J = \tfrac{p_{1,1}}{p_{1,0}+p_{1,1}} - \tfrac{p_{0,1}}{p_{0,0}+p_{0,1}}$ is Youden's index of $b$ on $\leaf$: its true
positive rate minus its false positive rate. Hence the minimizing $\hat p$ defers with positive probability exactly when
$J > \lambda$, that is, when the black box is informative enough on $\leaf$ compared to the cost of deferring; otherwise
$\hat p(\defer) = 0$ and the PIM-entropy reduces to the usual entropy $H(m)$. Intuitively, the criterion therefore
favors splits that separate regions where $b$ is informative from regions where it is not.

\emph{Proof sketch.} The proof proceeds in three steps. First, writing $s = \hat{p}(0)$ and $t = \hat{p}(1)$, so that $\hat{p}(\defer) = 1 - s - t$, straightforward computations show that the objective function equals $u(s) + v(t) + \lambda$, to be minimized over $s, t \geq 0$ with $s + t \leq 1$.
Next, we can compute directly the unconstrained solution $(s_{\star},t_{\star})$, which exists because each term of the objective either vanishes identically or tends to $+\infty$ at the boundary, and is unique by strict convexity.
Finally, we distinguish two cases according to the sign of $\Lambda$.
If $\Lambda > 0$, then $s_\star + t_\star < 1$, so $(s_\star, t_\star)$ is feasible and hence optimal: the leaf defers with positive probability.
If $\Lambda \leq 0$, a convexity argument shows that the constrained minimizer satisfies $s + t = 1$. Then the leaf never defers, and the minimum equals the entropy $H(m)$, i.e.\ the splitting criterion of standard \cart.
This argument assumes that certain sums of the $p_{i,j}$ are nonzero; the degenerate cases follow from the same steps with minor modifications.

\begin{proof}[Complete proof.]
We follow the steps outlined in the sketch above. First, writing $s = \hat{p}(0)$ and $t = \hat{p}(1)$, so that $\hat{p}(\defer) = 1 - s - t$, we have
\begin{align*}
  &\frac{1}{n_\leaf} \sum_{i : X_i \in \leaf} \lpimlog(X_i,Y_i,\hat{p})
  \\
  &\quad= - p_{1,1}\log\bigl(\hat{p}(1)+\hat{p}(\defer)\bigr)
    - p_{0,0}\log\bigl(\hat{p}(0)+\hat{p}(\defer)\bigr)
    - p_{1,0}\log \hat{p}(1)
    - p_{0,1}\log \hat{p}(0)
    + \lambda \hat{p}(\defer)
  \\
  &\quad= - p_{1,1}\log\bigl(1 - \hat{p}(0)\bigr)
    - p_{0,0}\log\bigl(1 - \hat{p}(1)\bigr)
    - p_{1,0}\log \hat{p}(1)
    - p_{0,1}\log \hat{p}(0)
    + \lambda \bigl(1-\hat{p}(0)-\hat{p}(1) \bigr)
  \\
  &\quad= - p_{1,1}\log\bigl(1-\hat{p}(0)\bigr) - p_{0,1}\log \hat{p}(0) - \lambda \hat{p}(0)
    - p_{0,0}\log\bigl(1-\hat{p}(1)\bigr) - p_{1,0}\log \hat{p}(1) - \lambda \hat{p}(1) + \lambda
  \\
  &\quad= u\bigl(\hat{p}(0)\bigr) + v\bigl(\hat{p}(1)\bigr) + \lambda.
\end{align*}
We denote the space of feasible solutions by $\mathcal{K}:=\{ (s,t) \in [0,1]^2 \mid s+t \leq 1 \}$. Consequently, the minimization problem becomes:
\begin{align*}
    \text{minimize}   \quad & F(s,t) := u(s) + v(t) \\
    \text{subject to} \quad & (s,t) \in \mathcal{K}.
\end{align*}
Next, observe that $u$ is strictly convex if $p_{0,1} + p_{1,1} > 0$, i.e.\ $b$ classifies at least one point of $\leaf$ with the label $1$, while otherwise $u(s) = -\lambda s$ is a decreasing linear function. Similarly, $v$ is strictly convex if $p_{0,0} + p_{1,0} > 0$, or equivalently $b$ predicts $0$ on at least one data point of $\leaf$, and $v(t) = -\lambda t$ otherwise. These properties will be used several times.
Furthermore, $F$ attains its minimum both on $[0,1]^2$ and on $\mathcal{K}$. Indeed, with the convention $0 \log 0 = 0$, each term $-p_{j,k}\log(\cdot)$ of $F$ either vanishes or tends to $+\infty$ as its argument tends to $0$, so $F$ is continuous with values in $(-\infty,+\infty]$. Since $F$ is finite in the interior, it attains its minimum on every compact set, in particular on $[0,1]^2$ and on $\mathcal{K}$. Moreover, the minimizer over $[0,1]^2$ is unique, since, by the properties above, $u$ and $v$ each have a unique minimizer on $[0,1]$. We then define:
\begin{align*}
  (s_{\star},t_{\star}) &= \argmin_{(s,t) \in [0,1]^2} F(s,t),
  \\
  (s_{\mathrm{opt}},t_{\mathrm{opt}}) &= \argmin_{(s,t) \in \mathcal{K}} F(s,t).
\end{align*}
Next, we show that either these two minimizers coincide or the minimizer over $\mathcal{K}$ lies on the diagonal $\{ (s,t) \in [0,1]^2 \mid s + t = 1 \}$. This follows from the fact that, as mentioned previously, $u$ and $v$ are strictly convex or decreasing linear functions. Indeed, suppose for contradiction that $(s_{\star},t_{\star}) \notin \mathcal{K}$ and $s_{\mathrm{opt}} + t_{\mathrm{opt}} < 1$. By continuity, there exists $\theta\in (0,1)$ such that the point
\begin{equation*}
    (s_\theta,t_\theta) = (1-\theta)(s_{\mathrm{opt}},t_{\mathrm{opt}}) + \theta(s_\star,t_\star)
\end{equation*}
satisfies $s_\theta + t_\theta < 1$.
Then, strict convexity gives the contradiction:
\begin{equation*}
    F(s_\theta,t_\theta) < (1-\theta) F(s_{\mathrm{opt}},t_{\mathrm{opt}}) + \theta F(s_\star,t_\star) \leq F(s_{\mathrm{opt}},t_{\mathrm{opt}}),
\end{equation*}
where the first inequality is strict because $(s_{\mathrm{opt}},t_{\mathrm{opt}}) \neq (s_\star,t_\star)$, one lying in $\mathcal{K}$ and the other not.
In the degenerate case in which $u$ or $v$ is linear, the conclusion is immediate, since $F$ is then
strictly decreasing in $s$ or $t$, and so any minimizer over $\mathcal{K}$ must satisfy
$s_{\mathrm{opt}} + t_{\mathrm{opt}} = 1$. We denote by $(s_{\dagger},t_{\dagger})$ the minimizer of $F$ restricted to the diagonal.

We then proceed to compute $(s_{\star}, t_{\star})$ and $(s_{\dagger},t_{\dagger})$ explicitly. Starting from $(s_{\star}, t_{\star})$, we first note that the derivatives of $u$ and $v$ can be written as
\begin{align*}
  u'(s)
  &= \frac{\lambda s^2 + (p_{0,1}+p_{1,1}-\lambda)s - p_{0,1}}{s(1-s)}
   = \frac{\varphi(s)}{s(1-s)},
  \\
  v'(t)
  &= \frac{\lambda t^2 + (p_{1,0}+p_{0,0}-\lambda)t - p_{1,0}}{t(1-t)}
   = \frac{\psi(t)}{t(1-t)}.
\end{align*}
  In particular, on $(0,1)$, $u'$ and $v'$ have the same sign as $\varphi$ and $\psi$, respectively. Now, $\varphi$ is an upward parabola with $\varphi(0) = -p_{0,1} \leq 0$ and $\varphi(1) = p_{1,1} \geq 0$, so its smaller root is non-positive and its larger root $r$ lies in $[0,1]$. Hence $\varphi < 0$ on $(0,r)$ and $\varphi > 0$ on $(r,1)$, so $u$ is decreasing on $[0,r]$ and increasing on $[r,1]$, which gives $s_\star = r$. The same argument applied to $\psi$, with $\psi(0) = -p_{1,0} \leq 0$ and $\psi(1) = p_{0,0} \geq 0$, shows that $t_\star$ is the larger root of $\psi$.

The computation of $(s_{\dagger},t_{\dagger})$ is simpler. Recalling that $m = p_{1,0}+p_{1,1}$, at a point $(s,1-s)$ of the diagonal we have
\begin{align*}
    F(s,1-s)
    &=
    -m \log(1-s) - (1-m) \log s - \lambda
    \\&=
    \mathrm{E}(m,1-s) - \lambda,
\end{align*}
where $\mathrm{E}(m,\cdot)$ is the binary cross-entropy, which is minimized when its second argument equals $m$, that is, at $s=1-m$. Hence $(s_{\dagger},t_{\dagger}) = (1-m,m)$, and $F(1-m,m) + \lambda = H(m)$, as stated in the lemma.

It remains to show that the condition that determines which of $(s_{\star},t_{\star})$ and $(s_{\dagger},t_{\dagger})$ is the minimizer over $\mathcal{K}$ is exactly $\Lambda \leq 0$.
First, note that:
\begin{align*}
    \varphi(1-m)
    &=
    \lambda (1-m)^2 + (p_{0,1}+p_{1,1}-\lambda)\,(1-m) - p_{0,1}
    \\&=
    \lambda (1-m)(1-m-1) +(p_{0,1}+p_{1,1})(p_{0,1} + p_{0,0} ) -  p_{0,1}
    \\&=
    -\lambda m(1-m) +(p_{0,1}+p_{1,1})(p_{0,1} + p_{0,0} ) -  p_{0,1}
    \\&=
    -\lambda m(1-m) + p_{0,1}^2 + p_{0,1}p_{0,0} + p_{1,1} p_{0,1} + p_{1,1}p_{0,0}  -  p_{0,1}
    \\&=
    -\lambda m(1-m)  + p_{1,1} p_{0,0} + p_{0,1} (p_{0,1} + p_{0,0}  + p_{1,1}   -  1)
    \\&=
    -\lambda m(1-m)  + p_{1,1} p_{0,0} - p_{0,1} p_{1,0}.
\end{align*}
Similarly, one can show that:
\begin{equation*}
    \psi(m)
    =
    -\lambda m(1-m)  + p_{1,1} p_{0,0} - p_{0,1} p_{1,0}.
\end{equation*}
This allows us to set
\begin{equation*}
    \Lambda= \varphi(1-m) = \psi(m) = -\lambda m(1-m)  + p_{1,1} p_{0,0} - p_{0,1} p_{1,0}.
\end{equation*}
Moreover, since the smaller roots of $\varphi$ and $\psi$ are non-positive, for every $x \in [0,1]$ we have $\varphi(x) > 0$ if and only if $x > s_{\star}$, and $\psi(x) > 0$ if and only if $x > t_{\star}$. Hence
\begin{equation*}
    \Lambda > 0
    \iff
    \begin{cases}
        \varphi(1-m) > 0   \\
        \psi(m) > 0 \\
    \end{cases}
    \iff
    \begin{cases}
        1-m > s_{\star}   \\
        m > t_{\star} \\
    \end{cases}
    \Rightarrow
    s_{\star} + t_{\star} < 1.
\end{equation*}
Similarly:
\begin{equation*}
    \Lambda \leq 0
    \iff
    \begin{cases}
        \varphi(1-m) \leq 0   \\
        \psi(m) \leq 0 \\
    \end{cases}
    \iff
    \begin{cases}
        1-m \leq s_{\star}   \\
        m \leq t_{\star} \\
    \end{cases}
    \Rightarrow
    s_{\star} + t_{\star} \geq 1.
\end{equation*}
In the first case, $(s_{\star},t_{\star})$ is the minimizer over $\mathcal{K}$, which gives $\pimentropy(\mathbf{p}) = u(s_\star) + v(t_\star) + \lambda$. In the second case, either $s_{\star} + t_{\star} > 1$, and the minimizer is $(s_{\dagger},t_{\dagger}) = (1-m,m)$ by the previous steps, or $s_{\star} + t_{\star} = 1$, which forces $(s_{\star},t_{\star}) = (1-m,m)$; in both cases $\pimentropy(\mathbf{p}) = H(m)$.
\end{proof}

\FloatBarrier

\section{Supplement to Section~\ref{sec:conversion_bound}}
\label{app:sec:bound}
\subsection{Further Background for Section~\ref{sec:conversion_bound}}
\label{app:conversion_bound}

\subsubsection{Details for the Example in Figure~\ref{fig:conversion_gap}}
\label{app:details_conversion_gap_figure}

\begin{figure}[htb]
  \centering
  \begin{tikzpicture}[>=stealth,
    level distance=1.6cm,
    level 1/.style={sibling distance=6.4cm},
    level 2/.style={sibling distance=3.2cm},
    edge from parent/.style={draw, black!70},
    split/.style={draw=black!70, rounded corners=2pt, inner sep=3pt,
                  font=\footnotesize},
    leaf/.style={draw=black!70, inner sep=3pt, align=center,
                 font=\footnotesize},
    elbl/.style={font=\scriptsize, black!70}]
  \node[split] {$x < \tfrac{7}{20}$}
    child { node[split] {$x < \tfrac15$}
      child { node[leaf] {$[0,\tfrac15)$\\predict $1$\\error rate $\tfrac14$}
              edge from parent node[elbl, left=2pt] {yes} }
      child { node[leaf] {$[\tfrac15,\tfrac{7}{20})$\\predict $0$\\error rate $\tfrac13$}
              edge from parent node[elbl, right=2pt] {no} }
      edge from parent node[elbl, left=2pt] {yes} }
    child { node[split] {$x < \tfrac35$}
      child { node[leaf] {$[\tfrac{7}{20},\tfrac35)$\\predict $0$\\error rate $\tfrac15$}
              edge from parent node[elbl, left=2pt] {yes} }
      child { node[leaf] {$[\tfrac35,1]$\\predict $1$\\error rate $\tfrac18$}
              edge from parent node[elbl, right=2pt] {no} }
      edge from parent node[elbl, right=2pt] {no} };
\end{tikzpicture}
  \caption{The tree $\hat \partition$ in the example of
    Figure~\ref{fig:conversion_gap}. Each leaf shows its interval, the
    majority label predicted there, and the error rate of that label.}
  \label{fig:tree_example}
\end{figure}
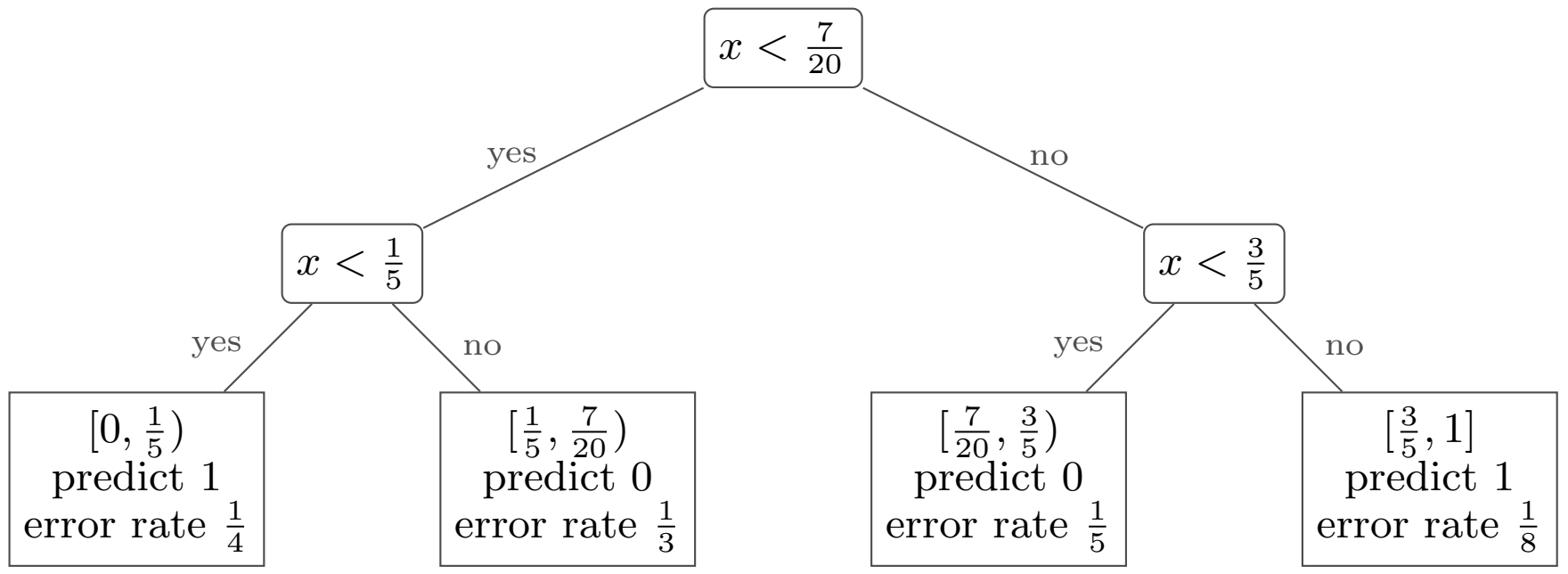

Here we provide the details for the example shown in
Figure~\ref{fig:conversion_gap}. The figure shows the case where
$\partitionclass$ corresponds to the partitions that can be generated by
decision trees of depth at most $2$. There is a single feature $X$ that
is uniformly distributed on $[0,1]$, and $Y = f^*(X)$ for the function
$f^*$ that alternates between $1$ and $0$ as shown in
Figure~\ref{fig:fstar_example}. The black box $b = f^*$ is perfect, so
$\alpha = 0$. We take the tree $\hat \partition$ to be as shown in
Figure~\ref{fig:tree_example}. We assume the conversion $\hat
\sigma_\lambda$ exactly minimizes the Lagrangian objective, so $\delta =
0$. This defers the leaves one at a time in decreasing order of their
error rates as $\lambda$ decreases.

\begin{figure}[htb]
  \centering
  \definecolor{cpblue}{HTML}{0072B2}
\begin{tikzpicture}[x=13cm, y=1.6cm, >=stealth,
    lbl/.style={font=\footnotesize},
    fstar/.style={cpblue, line width=1.2pt},
    jump/.style={cpblue!45, densely dotted, thin}]
  \draw[->, black!70] (0,-0.15) -- (1.04,-0.15) node[right, lbl, black] {$x$};
  \draw[->, black!70] (0,-0.15) -- (0,1.35) node[above, lbl, black] {$f^*(x)$};
  \foreach \y in {0,1} {
    \draw[black!70] (-0.006,\y) -- (0,\y);
    \node[left, lbl] at (-0.006,\y) {$\y$};
  }
  \foreach \x/\l in {0/$0$, 0.15/$\tfrac{3}{20}$, 0.2/$\tfrac15$,
                     0.25/$\tfrac14$, 0.35/$\tfrac{7}{20}$, 0.4/$\tfrac25$,
                     0.6/$\tfrac35$, 0.75/$\tfrac34$, 0.8/$\tfrac45$, 1/$1$} {
    \draw[black!70] (\x,-0.15) -- (\x,-0.21);
    \node[below, lbl] at (\x,-0.21) {\l};
  }
  \foreach \x in {0.15,0.2,0.25,0.35,0.4,0.6,0.75,0.8} {
    \draw[jump] (\x,0) -- (\x,1);
  }
  \draw[fstar] (0,1)    -- (0.15,1);
  \draw[fstar] (0.15,0) -- (0.2,0);
  \draw[fstar] (0.2,1)  -- (0.25,1);
  \draw[fstar] (0.25,0) -- (0.35,0);
  \draw[fstar] (0.35,1) -- (0.4,1);
  \draw[fstar] (0.4,0)  -- (0.6,0);
  \draw[fstar] (0.6,1)  -- (0.75,1);
  \draw[fstar] (0.75,0) -- (0.8,0);
  \draw[fstar] (0.8,1)  -- (1,1);
\end{tikzpicture}
  \caption{The function $f^*$ in the example of
    Figure~\ref{fig:conversion_gap}}
  \label{fig:fstar_example}
\end{figure}
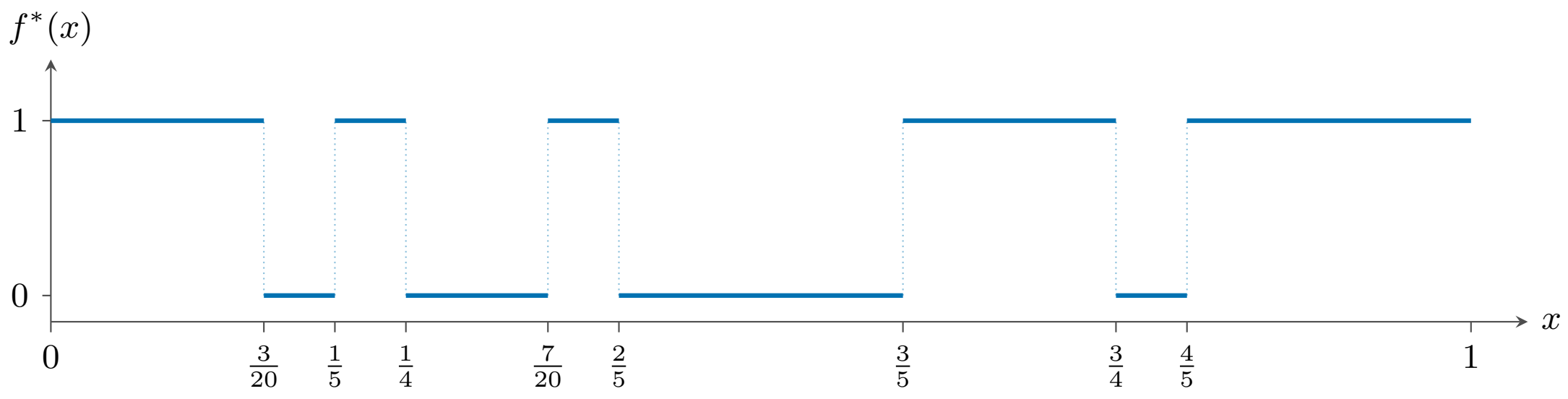

\subsubsection{Deferred Proofs from Section~\ref{sec:conversion_bound}}
\label{app:conversion_bound_proof_details}

\lemReduction*

\begin{proof}
  Recall the model class $\simpleG = \{(\hat \partition, \hat
  \sigma_\lambda) \mid \lambda \geq 0\}$ from the definition of
  $\gamma$.

  Now first suppose a subgradient $-\lambda$ of $\gamma$ exists at some
  $t$, and abbreviate $\hat f = (\hat \partition, \hat \sigma_\lambda)$.
  Then, by definition, $\big(U(\hat f),R(\hat f_{|b})\big) \in
  S(\simpleG)$ for $S$ as defined in \eqref{eqn:possible_U_R_values}.
  Taking $\hat t = U(\hat f)$, it follows that
  \[
    \gamma(\hat t) = \inf \{r \mid (u,r) \in \conv\big(S(\simpleG)\big),u\leq \hat t\}
               \leq R(\hat f_{|b}).
  \]
  Hence, by the definition of subgradients,
  \begin{align*}
    \gamma(t) -\lambda (\hat t - t) &\leq \gamma(\hat t)\\
    \gamma(t) +\lambda t
        &\leq \gamma(\hat t) + \lambda \hat t
        \leq R(\hat f_{|b}) + \lambda U(\hat f)
        = \Phi_\lambda(\hat f).
  \end{align*}
  In addition,
  \begin{align*}
    \gamma^*(t) + \lambda t
      &= \inf \{r \mid (u,r) \in \conv\big(S(\simpleF)\big),u\leq t\} +
      \lambda t\\
      &= \inf \{r + \lambda t \mid (u,r) \in
      \conv\big(S(\simpleF)\big),u\leq t\}\\
      &\geq \inf \{r + \lambda u \mid (u,r) \in
      \conv\big(S(\simpleF)\big),u\leq t\}\\
      &\geq \inf \{r + \lambda u \mid (u,r) \in
      \conv\big(S(\simpleF)\big)\}\\
      &= \inf \{r + \lambda u \mid (u,r) \in S(\simpleF)\}
      = \Phi^*_\lambda.
  \end{align*}
  Subtracting the two inequalities gives
  \[
    \gamma(t) - \gamma^*(t)
     = \gamma(t) + \lambda t - \big(\gamma^*(t) + \lambda t\big)
     \leq \Phi_\lambda(\hat f) - \Phi^*_\lambda,
  \]
  which is the conclusion of the lemma.

  It remains to establish existence of a subgradient $-\lambda$ of
  $\gamma$ at any $t$. To this end, observe that $\simpleG$ and hence
  $S(\simpleG)$ are finite, because there are only finitely many
  possible functions $\hat \sigma_\lambda$ from $\hat \partition$ to
  $\{0,1,\bot\}$. It follows that $K = \conv(S(\simpleG)) \subset
  [0,1]^2$ is a polytope.

  \newcommand{\umin}{u_\textrm{min}}
  \newcommand{\umax}{u_\textrm{max}}
  Let $[\umin,\umax] = \{u \mid \text{exists $r$ such that $(u,r) \in
  K$}\}$ be the interval obtained by projecting $K$ onto its first
  coordinate. Then, in fact, $\umin = 0$, which may be seen as follows:
  From \eqref{eqn:conversion_with_slack}, we have that
  \begin{equation}\label{eqn:zero_uncoverage}
    \lambda U\big((\hat \partition,\hat\sigma_\lambda)\big)
    \leq \Phi_\lambda\big((\hat \partition,\hat\sigma_\lambda)\big)
    \leq \min_\sigma \Phi_\lambda\big((\hat \partition,\sigma)\big) +
    \delta
    \leq 1 + \delta
    \qquad \text{for all $\lambda \geq 0$,}
  \end{equation}
  where the last inequality follows by taking $\sigma(\cell) = 0$, a
  function that never defers, and using that $R\big((\hat
  \partition,\sigma)_{|b}\big) \leq 1$. Since there are only finitely many
  possible functions $\hat \sigma_\lambda$, it follows that $U\big((\hat
  \partition,\hat\sigma_\lambda)\big)$ can only take finitely many
  different values when varying $\lambda$, and therefore
  \eqref{eqn:zero_uncoverage} implies that $U\big((\hat
  \partition,\hat\sigma_\lambda)\big) = 0$ for all sufficiently
  large~$\lambda$. Consequently, $K$ includes a point $(0,r)$ for some
  $r \in [0,1]$, which means that $\umin = 0$.

  Let
  \[
    g(u) = \min \{r : (u,r) \in K\}
  \]
  be the lower boundary of $K$, defined for $u \in [0,\umax]$. Then,
  since $K$ is a polytope, $g$ is convex and piecewise linear with a
  finite number of pieces. By definition,
  \[
    \gamma(t) = \min_{\substack{u \leq t\\u \in [0,\umax]}} g(u),
  \]
  which implies that $\gamma(t) = g(t)$ for $t \leq u_0$ and $\gamma(t)
  = g(u_0)$ for $t \geq u_0$, where $u_0 \in \argmin_{u \in [0,\umax]}
  g(u)$ is a minimizer of $g$. Consequently, $\gamma$ is also convex
  and piecewise linear with a finite number of pieces. It follows that
  $\gamma$ has at least one subgradient at any point $t \in [0,1]$,
  which equals the slope of a linear piece that contains~$t$. Since
  $\gamma$ is non-increasing, this slope will equal $-\lambda$ for some
  $\lambda \geq 0$, which remained to be shown.
\end{proof}

\lemDualGap*

\begin{proof}
  We start by showing that
  \[
    \Phi_\lambda\big((\hat \partition,\hat \sigma_\lambda)\big)
    \leq
    \min_\sigma \Phi_\lambda\big((\hat \partition,\sigma)\big)
    + \delta
    \leq
    \min\{R(\hat \partition),R(b) + \lambda\} + \delta.
  \]
  The first inequality is \eqref{eqn:conversion_with_slack}. The second follows
  because
  \[
    \min_\sigma \Phi_\lambda\big((\hat \partition,\sigma)\big)
    \leq
    \min_{\sigma : \hat \partition \to \{0,1\}} \Phi_\lambda\big((\hat \partition,\sigma)\big)
    = R(\hat \partition),
  \]
  and simultaneously we can upper bound the minimum by the $\sigma$
  that always defers, which gives
  \[
    \min_\sigma \Phi_\lambda\big((\hat \partition,\sigma)\big)
    \leq
    R(b) + \lambda.
  \]

  We proceed to find a lower bound for $\Phi^*_\lambda$. For
  readability, abbreviate
  \begin{align*}
    q_\cell &= \min\{P(Y = 0, X \in \cell), P(Y =
    1, X \in \cell)\}
            &
    e_\cell &= P(b(X) \neq Y, X \in \cell),
            &
    p_\cell &= P(X \in \cell).
  \end{align*}
  Then
  \begin{align*}
    \Phi^*_\lambda
      &= \inf_{\partition \in \partitionclass} \min_{\sigma : \partition
    \to \{0,1,\bot\}} \Phi_\lambda\big((\partition,\sigma)\big)\\
      &= \inf_{\partition \in \partitionclass}
        \sum_{\cell \in \partition} \min\{q_\cell, e_\cell + \lambda
        p_\cell\}\\
      &= \inf_{\partition \in \partitionclass}
        \sum_{\cell \in \partition} \min\big\{q_\cell,
        \big(R(b)+\lambda\big)p_\cell - \big(R(b)p_\cell - e_\cell\big)\big\}\\
      &\geq \inf_{\partition \in \partitionclass}
        \sum_{\cell \in \partition} \Big(\min\big\{q_\cell,
            \big(R(b)+\lambda\big)p_\cell\big\} - \big(R(b)p_\cell -
          e_\cell\big)_+\Big).
  \end{align*}
  Since $\sum_\cell \big(R(b)p_\cell - e_\cell\big) = R(b) - R(b) = 0$, we have
  \[
    \sum_\cell \big(R(b)p_\cell - e_\cell\big)_+
      = \sum_\cell \big(e_\cell - R(b)p_\cell\big)_+
      \leq \alpha.
  \]
  Furthermore,
  \[
    p_\cell = P(Y = 0, X \in \cell) + P(Y = 1, X \in \cell)
    \geq 2 q_\cell,
  \]
  which expresses that, even without deferring, the majority vote
  already achieves error at most $\frac{q_\cell}{p_\cell} \leq 1/2$. It
  follows that
  \[
    \inf_{\partition \in \partitionclass}
        \sum_{\cell \in \partition} \min\big\{q_\cell,
        \big(R(b)+\lambda\big)p_\cell\big\}
    \geq
    \inf_{\partition \in \partitionclass}
        \sum_{\cell \in \partition} q_\cell \min\big\{1,
        2\big(R(b)+\lambda\big)\big\}
    = R^* \min\big\{1, 2\big(R(b)+\lambda\big)\big\},
  \]
  which shows that the oracle's gain from deferring is at most a factor
  $\min\big\{1, 2\big(R(b)+\lambda\big)\big\}$ compared to the best
  possible labeling $R^*$ without deferring.
  Combining these inequalities, we find that
  \[
    \Phi^*_\lambda
    \geq R^* \min\big\{1, 2\big(R(b)+\lambda\big)\big\} - \alpha.
  \]
  All together,
  \begin{align*}
    \Phi_\lambda\big((\hat \partition,\hat \sigma_\lambda)\big)
    - \Phi^*_\lambda
    ~&\leq~
    \min\{R(\hat \partition),R(b) + \lambda\}
    - R^* \min\big\{1, 2\big(R(b)+\lambda\big)\big\} + \delta + \alpha\\
     &\leq~
     \max_{z \geq 0} \big(\min\{R(\hat \partition),z\}
    - R^* \min\{1, 2z\}\big) + \delta + \alpha\\
     &=~ R(\hat \partition)(1 - 2 R^*) + \delta + \alpha,
  \end{align*}
  where the last equality follows because the function is linear and
  non-decreasing in $z$ for $z \in [0,R(\hat \partition)]$, linear and
  non-increasing on $[R(\hat \partition),1/2]$ and constant on
  $[1/2,\infty)$, which implies that it is maximized at $z = R(\hat
  \partition)$. (We have used here that $0 \leq R^* \leq R(\hat
  \partition) = \sum_{\cell} q_\cell \leq \half \sum_{\cell} p_\cell =
  \half$.)
\end{proof}

\FloatBarrier

\section{Supplement to Section~\ref{sec:experiments}}
\label{app:sec:exp}
\subsection{Datasets}
\label{app:datasets}

Our experiments run on four families of data. Two are built from public tabular benchmarks used by the work we compare against, so that neither comparison is run on ground of our own choosing: the branch track of Section~\ref{sec:exp:trackb} uses a composite benchmark combining datasets from \citet{Chaou_R_B25BranchesEfficientlyseeking} and \citet{Frost_L_M_M24Partiallyinterpretablemodels}, and the conversion and decision-list tracks of Appendices~\ref{app:exp:convert} and~\ref{app:exp:frost} use that of \citet{Frost_L_M_M24Partiallyinterpretablemodels}. The third is a synthetic generator that
\emph{stipulates} the selective competence the method is supposed to exploit, and the fourth is
a concept bottleneck, where the interpretable features are themselves predicted.

All tabular data is public. The two tracks draw on the UCI repository, OpenML
\citep{Vansc_v_B_T14OpenMLnetworkedscience}, the LIBSVM collection
\citep{Chang_L11LIBSVMlibrarysupport},
\texttt{ustunb/risk-slim}, ProPublica's \texttt{compas-analysis}, the preprocessed binary
files distributed with \origbranches, and the dataset collection of
\citet{Lin_Z_H_R_S20Generalizedscalableoptimal}. The UCI datasets are released under CC BY 4.0
and the OpenML ones are marked public; the \texttt{risk-slim} data is BSD-3-Clause and the
\origbranches code MIT. CUB-200-2011 is restricted to non-commercial research and educational use.
ProPublica's COMPAS data and the FICO HELOC data come without a stated open licence and are
used here, as in prior work, for research only.

\begin{table}[!ht]
\centering\scriptsize
\setlength{\tabcolsep}{4pt}
\begin{tabular}{lrrrlrl}
\toprule
dataset & $n$ & $d$ & \%\,pos. & black box & acc. & task \\
\midrule
\multicolumn{7}{l}{\emph{both tracks}} \\
\texttt{diabetes} & 768 & 8 & 65.1 & \textsc{svm} & 76.0 & onset of diabetes in Pima Indian patients \\
\texttt{ficobin} & 10{,}459 & 17 & 47.8 & \textsc{gbt} & 73.0 & credit risk on a FICO home-equity line of credit \\
\texttt{heart\_statlog} & 270 & 13 & 44.4 & \textsc{svm} & 81.5 & presence of heart disease \\
\texttt{ionosphere} & 351 & 34 & 64.1 & \textsc{svm} & 94.4 & whether a radar return shows structure in the ionosphere \\
\texttt{mushroom} & 8{,}124 & 112 & 51.8 & \textsc{dt} & 100.0 & whether a mushroom is poisonous \\
\texttt{spambase} & 4{,}601 & 57 & 39.4 & \textsc{gbt} & 94.7 & whether an e-mail is spam \\
\midrule
\multicolumn{7}{l}{\emph{method comparison only}} \\
\texttt{banknote} & 1{,}372 & 4 & 44.5 & \textsc{rf} & 87.9 & genuine or forged banknote, from wavelet features of a scanned note \\
\texttt{breast\_cancer} & 286 & 32 & 29.7 & \textsc{rf} & 72.1 & recurrence of breast cancer within five years (Ljubljana) \\
\texttt{credit\_a} & 690 & 37 & 55.5 & \textsc{rf} & 84.1 & accept or decline a credit-card application \\
\texttt{german\_credit} & 1{,}000 & 48 & 70.0 & \textsc{rf} & 74.7 & good or bad credit risk for a loan applicant \\
\texttt{hepatitis} & 155 & 19 & 79.4 & \textsc{rf} & 80.9 & survival of a hepatitis patient \\
\texttt{monk1} & 124 & 17 & 50.0 & \textsc{rf} & 89.5 & first \textsc{monk} concept-learning problem (exact rule, noiseless) \\
\texttt{monk2} & 169 & 17 & 37.9 & \textsc{rf} & 58.8 & second \textsc{monk} problem (counting rule, hardest for a small tree) \\
\texttt{monk3} & 122 & 17 & 49.2 & \textsc{rf} & 100.0 & third \textsc{monk} problem (simple rule, 5\% label noise) \\
\texttt{pc1} & 1{,}109 & 21 & 6.9 & \textsc{rf} & 93.1 & defect-prone modules in NASA flight software \\
\texttt{sick} & 3{,}772 & 31 & 6.1 & \textsc{rf} & 92.6 & thyroid disease \\
\texttt{sonar} & 208 & 60 & 46.6 & \textsc{rf} & 82.5 & sonar return from a metal cylinder or a rock \\
\texttt{tictactoe} & 958 & 27 & 65.3 & \textsc{rf} & 96.5 & whether \texttt{x} has won, from a terminal board \\
\texttt{vote} & 435 & 16 & 38.6 & \textsc{rf} & 96.2 & a US congressman's party, from 16 recorded votes \\
\texttt{wdbc} & 569 & 30 & 37.3 & \textsc{rf} & 95.9 & malignant or benign breast mass, from cell-nucleus measurements \\
\midrule
\multicolumn{7}{l}{\emph{conversion and decision lists only}} \\
\texttt{adult} & 32{,}561 & 36 & 24.1 & \textsc{gbt} & 84.6 & whether a person earns more than \$50{,}000 \\
\texttt{bank} & 41{,}188 & 62 & 11.3 & \textsc{gbt} & 90.1 & whether a client subscribes after a marketing call \\
\texttt{bank2} & 41{,}188 & 63 & 11.3 & \textsc{gbt} & 92.4 & the same task, with the call-duration feature retained \\
\texttt{breastcancer} & 683 & 10 & 35.0 & \textsc{svm} & 96.4 & whether a tissue sample is benign \\
\texttt{careval} & 1{,}728 & 15 & --- & \textsc{nn} & 100.0 & car acceptability (four classes) \\
\texttt{compasbin} & 6{,}172 & 13 & 45.5 & \textsc{nn} & 68.8 & whether a convicted criminal re-offends \\
\texttt{mammo} & 961 & 14 & 46.3 & \textsc{nn} & 85.0 & whether a mammography sample is malignant \\
\bottomrule
\end{tabular}

\caption{Every dataset used in the paper, grouped by which experiments it appears in. $n$ is the
pooled size and $d$ the feature count; on the method comparison track the features are
binarized (four datasets ship binarized with the \origbranches distribution, the rest as described in the text), while the conversion and decision-list tracks keep them continuous
and scale them to $[0,1]$. \emph{black box} is the model the interpretable part defers to and
\emph{acc.} its held-out accuracy in points: a $100$-tree random forest on the datasets used for
the method comparison alone, and otherwise the class selected by the grid search of
\citet{Frost_L_M_M24Partiallyinterpretablemodels}. The six datasets in both tracks use one cached
black box and split throughout, so the two agree on every column here; the method comparison track calls
\texttt{heart} \texttt{heart\_statlog}. \%\,pos.\ is the proportion of the positive class over the
pooled data, and a dash marks \texttt{careval}, the one multi-class task. Sources and OpenML ids
are in \texttt{compute\_one.DATASET\_REGISTRY}.}
\label{tab:datasets}
\end{table}

\paragraph{Comparison of the three methods}
Section~\ref{sec:exp:trackb} relies on the $20$ datasets listed in Table~\ref{tab:datasets}. Numeric columns are binarized at their median and categorical ones are dummy-encoded (one indicator per category but the last); the medians are computed on the training split for the six datasets below and on the full dataset, which uses no labels, for the others. Each dataset is split once, so that all methods, including the expensive exact search, are compared on the same data. The split is a stratified $70/30$ at a fixed seed, and the black box is a $100$-tree random forest trained on the binarized features, except on the six datasets this
track shares with the conversion track --- \texttt{diabetes}, \texttt{ficobin},
\texttt{heart\_statlog}, \texttt{ionosphere}, \texttt{mushroom} and \texttt{spambase} ---
where the conversion track's
own cached black box and $80/20$ split are reused instead, so the two tracks defer to the same
black box wherever they overlap. For the greedy methods, the grown tree is pruned as in Section~\ref{sec:cart-convert}, with $\gamma = 0.01$, separately for each $\lambda$; Theorem~\ref{thm:one_eight_bound}, in contrast, analyzes the conversion of a single fixed partition.

The benchmark is deliberately heterogeneous in difficulty, and that matters for how the
results are read. Several datasets are degenerate for our purposes: the three \textsc{monk} problems are constructed, parity-like concepts on which greedy splitting is known to fail, so they favor exact search by construction, which is why Table~\ref{tab:trackb-summary} excludes them. Others are severely imbalanced
(\texttt{sick} at $6.1\%$ positive, \texttt{pc1} at $6.9\%$), where a constant predictor is
already strong and there is correspondingly little for deferral to buy.

Two of the twenty are listed for completeness but carry no result: exact search runs out of
memory on \texttt{mushroom} and \texttt{ficobin}, the two largest datasets of the track, so
neither enters the averages of Table~\ref{tab:trackb-summary}.
Appendix~\ref{app:trackb-tables} still reports the greedy criteria on them. Removing these two
and the three \textsc{monk} problems leaves the $15$ datasets over which
Section~\ref{sec:exp:trackb} averages.

\paragraph{Comparison with decision lists and cost of the Lagrangian relaxation}
A comparison between decision-list and decision tree is run in Appendix~\ref{app:exp:frost}, and an estimation of the
cost of the Lagrangian relaxation is done in Appendix~\ref{app:exp:convert}. Both use the same datasets
as the original comparison from \citet{Frost_L_M_M24Partiallyinterpretablemodels}. Those datasets and their description
are in Table~\ref{tab:datasets}.

\paragraph{Synthetic dataset}
Appendix~\ref{app:exp:synthetic} uses a synthetic dataset generator. Each draw is $2{,}000$
rows over $13$ binary features: an indicator $z$ and $12$ noisy copies of a hidden class $c$,
each flipped with probability $0.35$. The label is $c$ flipped with probability $0.25$. The
black box is correct wherever $z = 1$ and predicts at random where $z = 0$, with $z$
balanced and independent of $c$. Because $z$ is a feature, a tree \emph{can} isolate the
region where the black box is accurate. The split and tree settings are those of Section~\ref{sec:exp:trackb}, and results are
averaged over ten draws.

\paragraph{CUB-200}
Section~\ref{sec:exp:cbm} uses CUB-200-2011 \citep{Wah_B_W_P_B25CaltechUCSDbirds2002011}:
$11{,}788$ photographs of birds labeled with $200$ species, with the standard $112$-attribute
concept mask of \citet{Koh_N_T_M_P_K_L20Conceptbottleneckmodels}. We split it
$4{,}794 / 1{,}200 / 5{,}794$ into train, validation and test. Unlike every other dataset here
the interpretable features are not given: both the $112$ concepts and the $200$-way reference
model are predicted by separately fine-tuned Inception~v3 networks, so the tree reads
\emph{estimated} concepts and a path can be wrong about the bird even when its answer is right.
Following \citet{Havas_P_D22Addressingleakageconcept} the pipeline starts from a
$299 \times 299$ image. The baseline CBM's label head is a two-layer network with $200$ hidden
units, about $60$k parameters, against the $22.2$ million of the Inception~v3 that predicts the
species directly from pixels.

\paragraph{Computing infrastructure}
The greedy methods and the conversion and decision-list experiments ran on a laptop CPU (Intel
Core i7-10510U, 16 GB of RAM), and the CUB networks were fine-tuned on an NVIDIA RTX 5070 Ti
Laptop GPU (12 GB). \branches ran on the CPU nodes (AMD EPYC) of a national HPC cluster, with
$16$ cores and up to $28$ GB of memory per dataset, for about two hours each.

\FloatBarrier
\subsection{A Synthetic Dataset Where Deferral-Aware Splitting Does Help}
\label{app:exp:synthetic}
The comparison in Section~\ref{sec:exp:trackb} shows that on real data \pimcart and \branches do not outperform
\convert. This is consistent with Theorem~\ref{thm:one_eight_bound}, which bounds the gain of any tree learning
algorithm over conversion when the accuracy of the black box is homogeneous across the domain, something that we
observe in real data. We propose a synthetic dataset in which the accuracy of the black box is deliberately
inhomogeneous.

The dataset is generated as follows. A hidden class $c \in \{0,1\}$ is drawn uniformly, and
the tree sees $12$ binary features, each equal to $c$ flipped independently with probability
$0.35$, plus a balanced indicator $z$, independent of $c$. The label is $c$ flipped with
probability $0.25$. The black box is correct wherever $z = 1$ and predicts at random where
$z = 0$: on one half of the domain it has the information behind the label noise, which the
tree does not. Since $z$ carries no information about the label, a criterion that only looks
at labels has no reason to split on it, while a tree that does split on $z$ can defer exactly
where the black box is accurate. All methods use the protocol of
Section~\ref{sec:exp:trackb} (a $70/30$ split, maximum depth $10$, $\gamma = 0.01$, $20$
values of $\lambda$), and results are averaged over $10$ draws.

\begin{figure}[thb]
  \centering
  \includegraphics[width=0.5\linewidth]{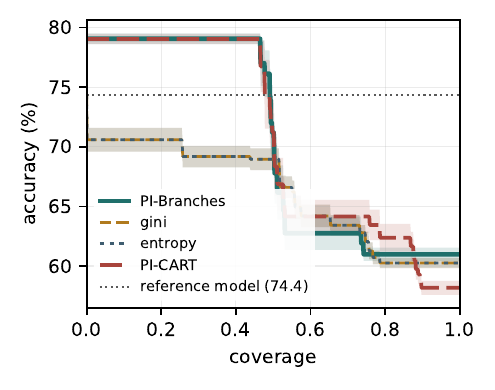}
  \caption{Coverage/accuracy on the held-out split of the synthetic dataset: best accuracy at
  coverage at least $c$, averaged over $10$ draws, with a band of one standard error. The
  dotted line is the black box alone.
\label{fig:synthetic-gadget}}
\end{figure}

As Figure~\ref{fig:synthetic-gadget} shows, both \branches and \pimcart are able to exploit the structure of the problem and significantly outperform
\convert. This means that deferral-aware splitting might be useful in some settings (even though we have not found them
in real data yet) and that our theorem correctly identifies that homogeneity of the black box accuracy drives the gap between
deferral-aware splitting and conversion.

\FloatBarrier
\subsection{Converting a Tree: How Much Does the Lagrangian Relaxation Cost?}
\label{app:exp:convert}

\begin{figure}[!ht]
  \centering
  \includegraphics[width=0.5\linewidth]{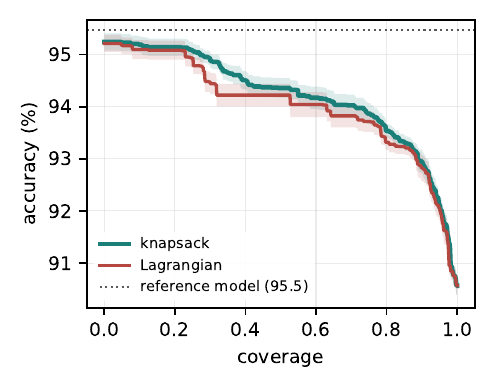}
  \caption{The two conversion rules on \texttt{spambase} at depth $6$, held-out split, mean over $10$ seeds with standard-error bands. The black box is refit on each seed's split, so its accuracy differs slightly from the single split of Table~\ref{tab:datasets}.}
  \label{fig:convert-rules}
\end{figure}

Figure~\ref{fig:pareto} shows that the Lagrangian relaxation can miss some optimal coverage/accuracy trade-offs. We
measure the difference between the Lagrangian relaxation and the exact solver on the conversion algorithm.
We compare two rules on the same fitted tree: the Lagrangian rule of
Section~\ref{sec:cart-convert}, which defers a leaf when the error rate of $b$ on it plus
$\lambda$ is below the error rate of its majority label, and the exact $0/1$ knapsack over
cells at each coverage budget. Each is fitted on the training split and applied unchanged to
the held-out split, over $13$ datasets at depths $2$ to $6$ with $10$ seeds.
Figure~\ref{fig:convert-rules} shows both rules on \texttt{spambase}, and
Table~\ref{tab:convert-rules} summarizes all datasets.

\begin{table}[!ht]
\centering\footnotesize
\setlength{\tabcolsep}{4pt}
\begin{tabular}{rrr}
\toprule
& \multicolumn{2}{c}{K $-$ L} \\
\cmidrule(lr){2-3}
depth & train & held out \\
\midrule
2 & $+0.265 \pm 0.088$ & $+0.268 \pm 0.086$ \\
3 & $+0.188 \pm 0.081$ & $+0.159 \pm 0.072$ \\
4 & $+0.145 \pm 0.059$ & $+0.177 \pm 0.055$ \\
5 & $+0.105 \pm 0.041$ & $+0.118 \pm 0.034$ \\
6 & $+0.069 \pm 0.024$ & $+0.086 \pm 0.022$ \\
\bottomrule
\end{tabular}

\caption{Converting one fitted tree into a partially interpretable model, 13 datasets $\times$ 10 seeds: gain of the
exact knapsack at each coverage budget (\textbf{K}) over the Lagrangian rule (\textbf{L}), both scored with the aggregate
score of Section~\ref{sec:exp:trackb}, in points, $\pm$ one standard error across the $13$ datasets. The difference is less than a percentage point.}
\label{tab:convert-rules}
\end{table}

\FloatBarrier
\subsection{An External Baseline: Partially Interpretable Decision Lists}
\label{app:exp:frost}

\citet{Frost_L_M_M24Partiallyinterpretablemodels} study the same predict-or-defer setting with a
different interpretable class: a \emph{decision list} that applies its rules in order and defers
whatever no rule covers. Their Appendix~C.5 compares that list against a decision-tree baseline
built the same way ours is --- fit the tree, then hand the weak leaves to the black box.

Their comparison reports \emph{one} operating point per method, the largest model whose
validation error stays within $0.005$ of the black box's, and tabulates its accuracy and its
coverage. Because the rule fixes an error budget rather than a coverage, the two methods are
read at \emph{different} coverages, and the two axes can disagree: on \texttt{spambase} their
list is slightly less accurate but covers considerably more. A single point also says nothing about the rest of either
method's frontier.

We therefore run the comparison on the whole frontier instead. Both classes are swept over the
full range of deferral budgets and scored \emph{at matched coverage}, so every comparison is
one-dimensional: at the same fraction of the data explained, which model is more accurate? We
are also deliberately generous to the baseline on both sides. The tree's hyperparameters come
from a cross-validated grid search, and its leaves defer by the
Lagrangian rule of Section~\ref{sec:cart-convert}; the list's own $(\ell, r^*)$ and rule count are
re-selected by cross-validation on every split.

\begin{figure}[!ht]
  \centering
  \includegraphics[width=0.5\linewidth]{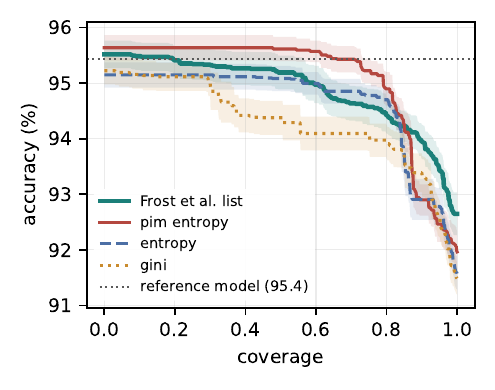}
  \caption{The decision list against the three tree criteria on \texttt{spambase}, over $10$
  redrawn splits; lines are mean envelopes and bands one standard error. Note the vertical scale: all four lie within $1.5$ points of each other across the whole range. The black box is refit on each split, so its accuracy differs slightly from the single split of Table~\ref{tab:datasets}.}
\label{fig:frost-lists}
\end{figure}

\begin{table}[!ht]
\centering\footnotesize
\setlength{\tabcolsep}{3pt}
\begin{tabular}{lrrrr}
\toprule
dataset & list & \multicolumn{3}{c}{list $-$ tree} \\
\cmidrule(lr){3-5}
 & & \texttt{gini} & \texttt{entropy} & \texttt{pim\_ent} \\
\midrule
adult & 83.85 & $+0.02$ & $+0.03$ & $+0.05$ \\
bank & 90.06 & $+0.01$ & $+0.01$ & $+0.02$ \\
bank2 & 91.34 & $-0.40$ & $-0.38$ & $-0.32$ \\
breastcancer & 95.03 & $+0.01$ & $+0.80$ & $-0.21$ \\
careval & 99.27 & $+1.27$ & $+1.34$ & --- \\
compasbin & 66.92 & $-0.77$ & $-0.98$ & $-0.94$ \\
diabetes & 72.99 & $-1.89$ & $-1.79$ & $-1.91$ \\
ficobin & 71.79 & $+0.05$ & $+0.00$ & $-0.04$ \\
heart & 76.85 & $-3.15$ & $-5.22$ & $-4.02$ \\
ionosphere & 90.50 & $-3.22$ & $-4.00$ & $-3.49$ \\
mammo & 80.31 & $-0.40$ & $-0.93$ & $-0.65$ \\
mushroom & 99.99 & $-0.01$ & $-0.01$ & $-0.01$ \\
spambase & 94.82 & $+0.51$ & $+0.16$ & $-0.04$ \\
\midrule
mean & & $-0.61$ & $-0.84$ & $-0.96$ \\
SE & & $\pm 0.37$ & $\pm 0.52$ & $\pm 0.41$ \\
\bottomrule
\end{tabular}

\caption{The partially interpretable \emph{decision lists} of \citet{Frost_L_M_M24Partiallyinterpretablemodels} against our trees, 13 datasets $\times$ 10 redrawn splits, in points. A \emph{negative} difference favors the tree. Every method is evaluated at the same set of deferral budgets per split, so the comparison does not reward whichever method happens to produce more candidate models. The average differences favor the trees but come from the small datasets, where the standard errors are large; on the three largest (\texttt{adult}, \texttt{bank}, \texttt{bank2}) every difference is at most $0.4$ points. \texttt{careval} has no \pimcart entry (the one multi-class dataset here).}
\label{tab:frost-lists}
\end{table}

Figure~\ref{fig:frost-lists} shows the frontiers on \texttt{spambase}, and
Table~\ref{tab:frost-lists} gives the head-to-head on every dataset of the benchmark, over
$10$ redrawn splits. Read this way the two classes come out close to
indistinguishable. On the four datasets large enough to resolve a difference ---
\texttt{adult}, \texttt{bank}, \texttt{bank2} and \texttt{ficobin}, from $10{,}459$ to
$41{,}188$ rows --- no comparison exceeds $0.4$ points in either direction, and on
\texttt{ficobin} none exceeds $0.05$. The aggregate differences of $-0.61$ to $-0.96$ points
are carried entirely by the small datasets, where the standard errors are $1.3$ to $1.8$ and
the per-dataset winner rotates.

Once both methods are tuned carefully and compared at matched coverage, we find no empirical
evidence of a difference between decision lists and trees. Appendix~C.5 of
\citet{Frost_L_M_M24Partiallyinterpretablemodels} reads each method at a single operating point,
where the comparison favors the list; over the whole frontier, we do not observe this
advantage.
\FloatBarrier
\subsection{Alternative Scoring Criteria}
\label{app:pimloss}

Every scoring in the main text summarizes the models a method produces over all values of $\lambda$. The aggregate score takes a
maximum over the method's own models at each coverage; the per-threshold columns select one
model per budget. Both therefore depend on \emph{which} models a method happens to produce,
and not only on how good they are. We report another scoring method in this appendix with no such dependence.

Each model was fitted to minimize a Lagrangian at a particular~$\lambda$, so we can evaluate each model on
exactly that quantity, at exactly that~$\lambda$:
\[
  L_\lambda(f) \;=\; R(f_{|b}) + \lambda\, U(f)
  \;=\; (1 - \text{accuracy}) + \lambda\,(1 - \text{coverage}).
\]
One model per method per~$\lambda$, compared against the others at the same~$\lambda$, with no
maximization and no selection. Lower is better. This is immune to the three artefacts that
affect the other scorings: unequal numbers of candidate models, the resolution of the
$\lambda$~grid, and padding where no value of $\lambda$ reaches full coverage.

We report our results in Table~\ref{tab:pimloss}.

The ordering differs from the aggregate score, and both differences are small.
We draw the same conclusion as in Section~\ref{sec:exp:trackb}: on this benchmark the four procedures perform very similarly.
Being a greedy surrogate, \pimcart does not minimize the Lagrangian objective directly, and on this benchmark it does not
reach a lower training value than \convert.
\begin{table}[!ht]
\centering\footnotesize
\setlength{\tabcolsep}{4pt}
\begin{tabular}{lrrrr}
\toprule
dataset & \texttt{gini} & \multicolumn{3}{c}{difference from \texttt{gini}} \\
\cmidrule(lr){3-5}
 & $L_\lambda$ & \branches & \texttt{ent.} & \texttt{pim\_ent.} \\
\midrule
banknote & 0.1233 & $+0.0000$ & $+0.0000$ & $+0.0000$ \\
breast\_cancer & 0.3229 & $+0.0003$ & $-0.0006$ & $+0.0232$ \\
credit\_a & 0.1945 & $+0.0000$ & $+0.0000$ & $+0.0000$ \\
diabetes & 0.2692 & $-0.0009$ & $+0.0000$ & $+0.0017$ \\
ficobin & 0.3019 & --- & $+0.0000$ & $-0.0023$ \\
german\_credit & 0.3449 & $+0.0000$ & $+0.0000$ & $+0.0270$ \\
heart\_statlog & 0.2989 & $-0.0017$ & $+0.0000$ & $-0.0096$ \\
hepatitis & 0.2065 & $+0.0214$ & $+0.0000$ & $+0.0397$ \\
ionosphere & 0.1464 & $+0.0370$ & $-0.0125$ & $+0.0378$ \\
monk1 & 0.0944 & $-0.0753$ & $+0.0000$ & $-0.0083$ \\
monk2 & 0.5485 & $-0.1429$ & $+0.0004$ & $-0.0188$ \\
monk3 & 0.0277 & $+0.0146$ & $+0.0000$ & $+0.0017$ \\
mushroom & 0.0132 & --- & $+0.0000$ & $+0.0031$ \\
pc1 & 0.0691 & $+0.0000$ & $+0.0000$ & $+0.0000$ \\
sick & 0.0616 & $+0.0000$ & $+0.0000$ & $+0.0000$ \\
sonar & 0.3278 & $-0.0337$ & $+0.0168$ & $-0.0215$ \\
spambase & 0.1885 & $+0.0020$ & $+0.0000$ & $+0.0000$ \\
tictactoe & 0.1260 & $+0.0168$ & $+0.0000$ & $+0.0520$ \\
vote & 0.0542 & $+0.0000$ & $+0.0000$ & $+0.0034$ \\
wdbc & 0.0582 & $+0.0104$ & $+0.0000$ & $+0.0068$ \\
\midrule
mean, 18 complete & 0.1924 & $-0.0084$ & $+0.0002$ & $+0.0075$ \\
mean, 18 complete (train) & 0.1138 & $-0.0115$ & $+0.0010$ & $+0.0061$ \\
mean, 15 ex-\textsc{monk} & 0.1861 & $+0.0034$ & $+0.0002$ & $+0.0107$ \\
mean, 15 ex-\textsc{monk} (train) & 0.1178 & $-0.0046$ & $+0.0002$ & $+0.0083$ \\
\bottomrule
\end{tabular}

\caption{PIM loss $L_\lambda = R + \lambda U$ on the 20 datasets, averaged over the $20$ values of~$\lambda$; \textbf{lower is better}.}
\label{tab:pimloss}
\end{table}
\FloatBarrier
\subsection{An Image the Rule Defers}
\label{app:bird-deferred}

\begin{figure}[!ht]
  \centering
  \includegraphics[width=\linewidth]{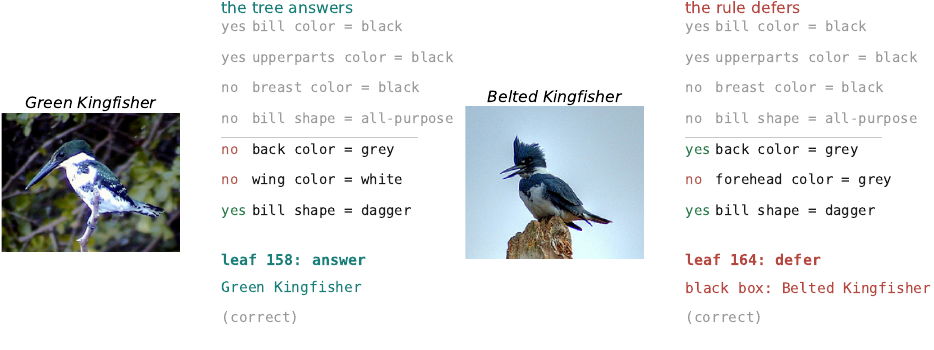}
  \caption{Output of the partially interpretable decision tree on two Kingfisher birds. The tree predicts the class of
  the bird on the left and defers the prediction of the bird on the right to the blackbox model.}
  \label{fig:bird_deferred}
\end{figure}

In Figure~\ref{fig:bird_explanation}, we present an image of a bird that is not deferred by our depth-$8$ tree. In
Figure~\ref{fig:bird_deferred}, we additionally provide an image of a similar bird on which the tree will defer: their concepts agree on the first four nodes of the tree, and they diverge at
\emph{back color = grey}.

\FloatBarrier
\subsection{Per-Dataset Results}
\label{app:trackb-tables}
We report average scores over datasets in Table~\ref{tab:trackb-summary}. Table~\ref{tab:trackb-full-A} gives the scores for each dataset.
Since \branches performs exact search, it gives us a lower bound on the training objective wherever it proves
optimality, which it does for every $\lambda$ on $13$ of the $18$ datasets on which it terminates; the \textbf{opt.}
column of Table~\ref{tab:trackb-full-A} counts, for each dataset, the values of $\lambda$ on which optimality was proved.

\begin{table*}[!ht]
\centering\scriptsize
\renewcommand{\arraystretch}{0.8}
\setlength{\tabcolsep}{4pt}
\begin{tabular}{llrrrrrrrr}
\toprule
dataset & method & opt. & agg.\ score & \multicolumn{2}{c}{$c \geq 0.25$} & \multicolumn{2}{c}{$c \geq 0.50$} & \multicolumn{2}{c}{$c \geq 0.75$} \\
\cmidrule(lr){5-6}
\cmidrule(lr){7-8}
\cmidrule(lr){9-10}
 & & & & acc & cov & acc & cov & acc & cov \\
\midrule
banknote & \branches & \textbf{20/20} & \textbf{87.9} & 87.6 & 79.4 & 87.6 & 79.4 & 87.6 & 79.4 \\
 & \texttt{gini} &  & \textbf{87.9} & 87.6 & 79.4 & 87.6 & 79.4 & 87.6 & 79.4 \\
 & \texttt{entropy} &  & \textbf{87.9} & 87.6 & 79.4 & 87.6 & 79.4 & 87.6 & 79.4 \\
 & \pimcart &  & \textbf{87.9} & 87.6 & 79.4 & 87.6 & 79.4 & 87.6 & 79.4 \\
\addlinespace[1pt]
breast\_cancer & \branches & 3/20 & 75.0 & 72.1 & 17.4 & 75.6 & 60.5 & 74.4 & 87.2 \\
 & \texttt{gini} &  & \textbf{75.2} & 72.1 & 43.0 & 75.6 & 74.4 & 74.4 & 94.2 \\
 & \texttt{entropy} &  & \textbf{75.2} & 72.1 & 48.8 & 75.6 & 74.4 & 74.4 & 94.2 \\
 & \pimcart &  & \textbf{75.2} & 72.1 & 32.6 & 75.6 & 74.4 & 74.4 & 94.2 \\
\addlinespace[1pt]
credit\_a & \branches & 4/20 & \textbf{84.0} & 84.5 & 47.3 & 84.5 & 78.3 & 84.5 & 78.3 \\
 & \texttt{gini} &  & \textbf{84.0} & 84.5 & 47.3 & 84.5 & 78.3 & 84.5 & 78.3 \\
 & \texttt{entropy} &  & \textbf{84.0} & 84.5 & 47.3 & 84.5 & 78.3 & 84.5 & 78.3 \\
 & \pimcart &  & \textbf{84.0} & 84.5 & 47.3 & 84.5 & 78.3 & 84.5 & 78.3 \\
\addlinespace[1pt]
diabetes & \branches & \textbf{20/20} & \textbf{76.7} & 76.6 & 64.9 & 76.6 & 64.9 & 77.3 & 87.0 \\
 & \texttt{gini} &  & 75.5 & 76.6 & 55.2 & 74.7 & 89.0 & 74.7 & 89.0 \\
 & \texttt{entropy} &  & 75.5 & 76.6 & 55.2 & 74.7 & 89.0 & 74.7 & 89.0 \\
 & \pimcart &  & 75.7 & 76.6 & 64.9 & 76.6 & 64.9 & 74.7 & 89.0 \\
\addlinespace[1pt]
ficobin & \branches &  & --- & --- & --- & --- & --- & --- & --- \\
 & \texttt{gini} &  & 69.6 & 69.6 & 100.0 & 69.6 & 100.0 & 69.6 & 100.0 \\
 & \texttt{entropy} &  & 69.6 & 69.6 & 100.0 & 69.6 & 100.0 & 69.6 & 100.0 \\
 & \pimcart &  & \textbf{71.9} & 71.7 & 82.4 & 71.7 & 82.4 & 71.7 & 82.4 \\
\addlinespace[1pt]
german\_credit & \branches & 4/20 & \textbf{74.6} & 75.7 & 45.7 & 73.7 & 75.7 & 73.7 & 100.0 \\
 & \texttt{gini} &  & \textbf{74.6} & 75.7 & 45.7 & 73.7 & 75.7 & 73.7 & 100.0 \\
 & \texttt{entropy} &  & \textbf{74.6} & 75.7 & 45.7 & 73.7 & 75.7 & 73.7 & 100.0 \\
 & \pimcart &  & 73.4 & 74.3 & 26.3 & 74.7 & 65.3 & 70.0 & 100.0 \\
\addlinespace[1pt]
heart\_statlog & \branches & \textbf{20/20} & \textbf{78.2} & 83.3 & 20.4 & 74.1 & 87.0 & 74.1 & 87.0 \\
 & \texttt{gini} &  & 73.5 & 75.9 & 27.8 & 75.9 & 48.1 & 72.2 & 85.2 \\
 & \texttt{entropy} &  & 73.5 & 75.9 & 27.8 & 75.9 & 48.1 & 72.2 & 85.2 \\
 & \pimcart &  & 77.1 & 75.9 & 51.9 & 75.9 & 51.9 & 72.2 & 81.5 \\
\addlinespace[1pt]
hepatitis & \branches & \textbf{20/20} & \textbf{83.5} & 85.1 & 66.0 & 85.1 & 66.0 & 80.9 & 95.7 \\
 & \texttt{gini} &  & 82.7 & 80.9 & 42.6 & 80.9 & 63.8 & 80.9 & 97.9 \\
 & \texttt{entropy} &  & 82.7 & 80.9 & 42.6 & 80.9 & 63.8 & 80.9 & 97.9 \\
 & \pimcart &  & 80.9 & 78.7 & 59.6 & 78.7 & 59.6 & 74.5 & 100.0 \\
\addlinespace[1pt]
ionosphere & \branches & 7/20 & 88.0 & 87.3 & 52.1 & 87.3 & 52.1 & 90.1 & 80.3 \\
 & \texttt{gini} &  & 90.3 & 91.5 & 54.9 & 91.5 & 54.9 & 88.7 & 94.4 \\
 & \texttt{entropy} &  & 90.4 & 91.5 & 54.9 & 91.5 & 54.9 & 88.7 & 93.0 \\
 & \pimcart &  & \textbf{91.8} & 88.7 & 35.2 & 84.5 & 66.2 & 91.5 & 100.0 \\
\addlinespace[1pt]
monk1 & \branches & \textbf{20/20} & \textbf{100.0} & 100.0 & 100.0 & 100.0 & 100.0 & 100.0 & 100.0 \\
 & \texttt{gini} &  & 94.1 & 92.1 & 89.5 & 92.1 & 89.5 & 92.1 & 100.0 \\
 & \texttt{entropy} &  & 94.1 & 92.1 & 89.5 & 92.1 & 89.5 & 92.1 & 100.0 \\
 & \pimcart &  & 94.7 & 94.7 & 100.0 & 94.7 & 100.0 & 94.7 & 100.0 \\
\addlinespace[1pt]
monk2 & \branches & \textbf{20/20} & \textbf{72.5} & 62.7 & 68.6 & 62.7 & 68.6 & 72.5 & 92.2 \\
 & \texttt{gini} &  & 51.6 & 56.9 & 29.4 & 49.0 & 86.3 & 49.0 & 86.3 \\
 & \texttt{entropy} &  & 54.3 & 56.9 & 29.4 & 56.9 & 62.7 & 56.9 & 62.7 \\
 & \pimcart &  & 70.6 & 58.8 & 58.8 & 58.8 & 58.8 & 70.6 & 100.0 \\
\addlinespace[1pt]
monk3 & \branches & \textbf{20/20} & 99.4 & 100.0 & 56.8 & 100.0 & 56.8 & 100.0 & 81.1 \\
 & \texttt{gini} &  & \textbf{99.8} & 100.0 & 56.8 & 100.0 & 56.8 & 100.0 & 91.9 \\
 & \texttt{entropy} &  & \textbf{99.8} & 100.0 & 56.8 & 100.0 & 56.8 & 100.0 & 91.9 \\
 & \pimcart &  & 99.5 & 100.0 & 81.1 & 100.0 & 81.1 & 100.0 & 81.1 \\
\addlinespace[1pt]
mushroom & \branches &  & --- & --- & --- & --- & --- & --- & --- \\
 & \texttt{gini} &  & 98.9 & 98.9 & 100.0 & 98.9 & 100.0 & 98.9 & 100.0 \\
 & \texttt{entropy} &  & 98.9 & 98.9 & 100.0 & 98.9 & 100.0 & 98.9 & 100.0 \\
 & \pimcart &  & \textbf{99.7} & 100.0 & 72.9 & 100.0 & 72.9 & 98.6 & 91.1 \\
\addlinespace[1pt]
pc1 & \branches & \textbf{20/20} & \textbf{93.1} & 93.1 & 100.0 & 93.1 & 100.0 & 93.1 & 100.0 \\
 & \texttt{gini} &  & \textbf{93.1} & 93.1 & 100.0 & 93.1 & 100.0 & 93.1 & 100.0 \\
 & \texttt{entropy} &  & \textbf{93.1} & 93.1 & 100.0 & 93.1 & 100.0 & 93.1 & 100.0 \\
 & \pimcart &  & \textbf{93.1} & 93.1 & 100.0 & 93.1 & 100.0 & 93.1 & 100.0 \\
\addlinespace[1pt]
sick & \branches & \textbf{20/20} & \textbf{93.9} & 93.9 & 100.0 & 93.9 & 100.0 & 93.9 & 100.0 \\
 & \texttt{gini} &  & \textbf{93.9} & 93.9 & 100.0 & 93.9 & 100.0 & 93.9 & 100.0 \\
 & \texttt{entropy} &  & \textbf{93.9} & 93.9 & 100.0 & 93.9 & 100.0 & 93.9 & 100.0 \\
 & \pimcart &  & \textbf{93.9} & 93.9 & 100.0 & 93.9 & 100.0 & 93.9 & 100.0 \\
\addlinespace[1pt]
sonar & \branches & 7/20 & 74.1 & 69.8 & 79.4 & 69.8 & 79.4 & 69.8 & 79.4 \\
 & \texttt{gini} &  & 66.9 & 66.7 & 61.9 & 66.7 & 61.9 & 65.1 & 100.0 \\
 & \texttt{entropy} &  & 67.0 & 68.3 & 74.6 & 68.3 & 74.6 & 65.1 & 84.1 \\
 & \pimcart &  & \textbf{77.9} & 73.0 & 66.7 & 73.0 & 66.7 & 66.7 & 96.8 \\
\addlinespace[1pt]
spambase & \branches & \textbf{20/20} & 87.0 & 93.6 & 30.5 & 90.6 & 52.8 & 83.9 & 75.5 \\
 & \texttt{gini} &  & \textbf{87.5} & 93.6 & 30.5 & 90.9 & 53.7 & 83.0 & 79.8 \\
 & \texttt{entropy} &  & \textbf{87.5} & 93.6 & 30.5 & 90.9 & 53.7 & 83.0 & 79.8 \\
 & \pimcart &  & \textbf{87.5} & 93.6 & 30.5 & 90.9 & 53.7 & 83.0 & 79.8 \\
\addlinespace[1pt]
tictactoe & \branches & \textbf{20/20} & 91.4 & 95.8 & 53.1 & 95.8 & 53.1 & 86.5 & 94.4 \\
 & \texttt{gini} &  & \textbf{95.1} & 95.5 & 40.6 & 95.5 & 52.4 & 95.1 & 84.0 \\
 & \texttt{entropy} &  & \textbf{95.1} & 95.5 & 40.6 & 95.5 & 52.4 & 95.1 & 84.0 \\
 & \pimcart &  & 90.7 & 92.4 & 39.6 & 87.5 & 100.0 & 87.5 & 100.0 \\
\addlinespace[1pt]
vote & \branches & \textbf{20/20} & \textbf{96.1} & 96.2 & 58.0 & 96.2 & 58.0 & 96.2 & 85.5 \\
 & \texttt{gini} &  & \textbf{96.1} & 96.2 & 58.0 & 96.2 & 58.0 & 96.2 & 85.5 \\
 & \texttt{entropy} &  & \textbf{96.1} & 96.2 & 58.0 & 96.2 & 58.0 & 96.2 & 85.5 \\
 & \pimcart &  & \textbf{96.1} & 96.2 & 58.0 & 96.2 & 58.0 & 96.2 & 85.5 \\
\addlinespace[1pt]
wdbc & \branches & \textbf{20/20} & \textbf{97.4} & 96.5 & 49.1 & 96.5 & 49.1 & 95.9 & 88.3 \\
 & \texttt{gini} &  & 96.4 & 96.5 & 49.1 & 96.5 & 49.1 & 95.9 & 88.3 \\
 & \texttt{entropy} &  & 96.4 & 96.5 & 49.1 & 96.5 & 49.1 & 95.9 & 88.3 \\
 & \pimcart &  & 96.5 & 96.5 & 49.1 & 96.5 & 49.1 & 95.3 & 79.5 \\
\addlinespace[1pt]
\bottomrule
\end{tabular}

\caption{Score of \branches, \convert (\texttt{gini} and \texttt{entropy}) and \pimcart by dataset. Columns as in Table~\ref{tab:trackb-summary};
\textbf{agg.\ score} is the mean of the coverage/accuracy staircase over $[0,1]$ and the best method on each dataset is
in bold. \textbf{opt.} counts the values of $\lambda$ on which \branches proved optimality. A dash marks \texttt{mushroom} and \texttt{ficobin}, where exact search did not finish; the greedy criteria are still reported there. Under $c \geq t$ the model with the best training accuracy among those covering at least $t$ is selected: \textbf{cov} is the coverage it achieves on the held-out split, often well above $t$ because none of the values of $\lambda$ yields a model near it.}
\label{tab:trackb-full-A}
\end{table*}
\FloatBarrier

\end{document}